\documentclass{bmvc2k}

\title{TASTE: A Designer-Annotated Multi-Dimensional Preference Dataset for
       AI-Generated Graphic Design\thanks{$^{*}$Equal contribution. $^{\dagger}$Corresponding author.}}

\addauthor{Haonan Zhu$^{*}$}{haonan@lica.world}{1}
\addauthor{Elad Hirsch$^{*}$}{elad@lica.world}{1}
\addauthor{Alexandria Minetti}{alli.minetti@contra.com}{2}
\addauthor{Allison Nulty}{allison.nulty@contra.com}{2}
\addauthor{Ben Huffman}{ben@contra.com}{2}
\addauthor{Purvanshi Mehta$^{\dagger}$}{purvanshi@lica.world}{1}

\addinstitution{Lica World, San Francisco, USA.}
\addinstitution{Contra.Work Inc., Brooklyn, NY, USA.}

\runninghead{Zhu et al.}{TASTE: Designer-Annotated Preference for AI Design}

\def\ie{\emph{i.e}\bmvaOneDot}

\usepackage{booktabs}
\usepackage{amssymb}
\usepackage{amsmath}
\usepackage{placeins}
\usepackage{etoolbox}

\newcommand{\pmax}{p_{\max}}

\usepackage{geometry}
\makeatletter
\patchcmd{\bmv@maketitle}{\newpage}{\newpage\vspace*{1cm}}{}{}
\makeatother

\usepackage{hyperref}
\hypersetup{colorlinks=true, urlcolor=blue, linkcolor=blue, citecolor=blue}

\begin{document}

\maketitle

\begin{abstract}

Text-to-image models now generate graphic design at production scale, yet their supervision still comes primarily from photo-style preference datasets with a single overall verdict per comparison. Designers evaluate designs along several distinct axes (e.g., typography, layout, color harmony) that a single preference label collapses.  We release \emph{TASTE} \textit{(Typography, Aesthetics, Spatial, Tone, Etc.)}, a multi-dimensional preference dataset in which two disjoint cohorts of five professional designers each ranked outputs from four current text-to-image models across nine criteria along with per-image hallucination flags. We pair the dataset with two contributions. First, a criterion-agnostic signal-validation framework based on Kendall's $\tau$, majority-vote probability, and Condorcet cycles against exact iid-uniform nulls; the analysis reveals significant but moderate designer agreement, with every TASTE criterion rejecting the random-rater null. Second, we benchmark preference models on TASTE and find that off-the-shelf VLM judges and dedicated T2I scorers fail to reach majority agreement with the designer panel, while a small MLP head trained directly on TASTE substantially narrows the gap to the single-rater ceiling, setting a baseline for future TASTE-trained preference models.

\end{abstract}

\section{Introduction}
\label{sec:intro}

Text-to-image (T2I) models have crossed from research artifact to deployed design tool. Current systems generate UI mockups, posters, ad creatives, and typographic compositions end-to-end and ship them to end users in production design pipelines; a recent survey of AI-driven graphic design~\cite{Survey2025DesignAI} catalogues the scope of this shift. 
Yet these systems are still largely evaluated and optimized using preference signals derived from photographic generation. This creates a misalignment between what models are rewarded for and what downstream design users actually value, potentially favoring photorealistic plausibility over typography, hierarchy, or structural fidelity.

The public preference datasets that drive evaluation, reward
modeling and preference optimization are collected on
photo-style generation rather than on design.  HPSv2 and HPSv3
\cite{Wu2023HPSv2, Ma2025HPSv3}, Pick-a-Pic~\cite{Kirstain2023PickaPic},
ImageReward~\cite{Xu2023ImageReward}, RichHF-18K~\cite{Liang2024RichHF},
MHP/MPS~\cite{Zhang2024MPS}, and VisionReward~\cite{Xu2026VisionReward}
all collect preferences over photo-style outputs.  Survey
work~\cite{Survey2025DesignAI} confirms the gap, and
DesignPref~\cite{Peng2025DesignPref} describes a UI-design pairwise
dataset whose data has not yet been released.  A complementary line
of recent work targets the structural and task-level side of design:
LICA~\cite{Hirsch2026LICA} provides 1.55M layered design compositions
with structural metadata, and
Graphic-Design-Bench~\cite{Deganutti2026GDB} evaluates current models
on 50 design tasks across layout, typography, infographics, template
semantics, and animation.  These resources test what models
\emph{can do}; neither records what designers \emph{prefer}.

Sub-dimensional preference remains a missing piece.  
Photo-style
preference is dominated by a small set of objective cues such as
anatomical plausibility and prompt alignment, so a single ``which is
better'' label captures most of the signal.  Graphic design splits
along several partly independent axes including typography rules,
visual hierarchy, color harmony, mood and tone, spatial layout
fidelity, and description-fidelity criteria of the form \emph{does
the color requested in the prompt actually appear in the output}.
A single overall label averages over these axes and loses the
per-axis signal.  
A layout may preserve spatial structure while violating color intent; another may satisfy prompt semantics but break typographic hierarchy. Both could receive the same overall preference label despite failing for different reasons.
Work on visual importance and saliency in graphic
design supports treating these axes as distinct cognitive
processes~\cite{Bylinskii2017VisualImportance, Fosco2020Predimportance},
and multi-axis benchmarking is standard practice in adjacent
modalities, for example VBench and VBench-2.0 in
video~\cite{Huang2024VBench, Huang2025VBench2}.

Building a design-preference dataset is not sufficient on its own: the ratings must encode stable, learnable signal rather than arbitrary ratings. Similar aggregate disagreement can reflect random ratings, a weak shared preference with rater-specific noise, or designer groups with internally consistent but incompatible tastes. We propose a three-part diagnostic for separating these cases: graded agreement (per-prompt Kendall's $\tau$), majority decisiveness (majority-vote probability $\pmax$), and triadic intransitivity (Condorcet cycle existence). Tested against exact null distributions at $(p=4, R=5)$, every TASTE criterion rejects the random-rater null on at least two of the three statistics. Three subsampled
reference datasets at the same shape (Sushi~\cite{Kamishima2003Sushi},
MovieLens~\cite{Harper2015MovieLens}, and the HPSv2 test data~\cite{Wu2023HPSv2}) place designer agreement on graphic design between food and movie preferences and below photo-style image-quality preference. Signal level varies across
criteria, with typography and spatial accuracy carrying roughly twice the signal of color harmony.  Beyond TASTE, we argue that this kind of analysis should be standard practice on any new preference dataset before model training begins, since it answers the prior question of whether learnable signal is present.

Off-the-shelf systems do not yet read this signal.  Across six
open-weight VLM judges spanning four families and three dedicated
T2I-preference scorers (HPSv2.1, PickScore-v1, LAION-Aesthetic-V2),
no pre-trained system exceeds $0.55$ macro agreement with the
5-designer majority on TASTE under a position-bias-corrected
protocol~\cite{Zheng2023MTBench}.  Within the VLM slate,
position-bias rate and content sensitivity trade off (Spearman
$\rho{=}{+}0.94$), so scaling moves the operating point along this
frontier rather than to a better one.

TASTE is designed not only as a benchmark, but as a decision layer for design generation systems. Its criterion-level structure enables fine-grained evaluation of model strengths and weaknesses, allowing practitioners to select or route among generators based on task-specific priorities (e.g., typography fidelity, spatial accuracy, or color control) rather than relying on a single aggregate preference score. This same structure provides supervision for training preference judges, reward models, and alignment objectives that optimize for specific design dimensions, enabling more controllable and designer-aligned generation.

Hence, our work makes the following contributions:

\begin{enumerate}
    \item We release \emph{TASTE}\footnote{Code: \url{https://github.com/purvanshi-lica/taste}.  Data: \url{https://huggingface.co/datasets/purvanshi/TASTE}.} \textit{(Typography, Aesthetics, Spatial, Tone, Etc.)}, a designer-annotated preference dataset for AI-generated graphic design recording one rating per criterion across nine criteria, with ten professional designers ranking outputs from four current T2I models (1,600 ratings per criterion, plus hallucination flags on overall-preference cohorts). Both pairwise comparisons and derived 4-way rankings are released.
    \item We introduce a criterion-agnostic diagnostic framework for validating preference-signal quality in new datasets, based on graded agreement, majority decisiveness, and triadic intransitivity.  Applied with cross-domain anchors from Sushi, MovieLens, and HPSv2-test at the same shape, every TASTE criterion rejects the random-rater null and designer agreement on graphic design sits between subjective-taste and photo-quality regimes.
    \item We benchmark preference models on TASTE.  Nine pre-trained systems, three dedicated T2I-preference scorers (HPSv2.1, PickScore-v1, LAION-Aesthetic-V2) and six open-weight VLMs prompted as judges across four families and an order-of-magnitude range in active parameter count, all fall below $0.55$ macro agreement with the 5-designer majority, and scaling within the VLM slate does not close the gap.  A small MLP head trained directly on TASTE reaches $0.611$ pairwise accuracy, approaching the $0.741$ single-rater ceiling and setting a baseline for future TASTE-trained preference models.
\end{enumerate}

\section{Related work}
\label{sec:related}

\paragraph{Graphic-design ecosystem.}
LICA~\cite{Hirsch2026LICA} releases 1.55M layered design compositions
with structural metadata covering text, image, vector, and group
elements.  Graphic-Design-Bench~\cite{Deganutti2026GDB} builds on LICA to evaluate current models on 50 understanding-and-generation tasks
across layout, typography, infographics, template semantics, and
animation.  These two resources cover the structural and task sides
of graphic-design AI but record no human preferences.  A recent
survey~\cite{Survey2025DesignAI} maps the broader landscape and
flags multi-rater design preference data as missing.  
Other works \cite{Bylinskii2017VisualImportance, Fosco2020Predimportance} provide
foundational work on visual-importance prediction across graphic
design types and motivate sub-dimensional analysis of design
quality.  TASTE positions on top of this stack as the human-preference
layer.

\paragraph{Design-domain preference.}
DesignPref~\cite{Peng2025DesignPref} is the closest in spirit to
TASTE.  It collects 12k pairwise UI design comparisons from 20
designers with identity-linked annotations and reports Krippendorff
$\alpha = 0.25$ on a single overall preference label.  TASTE differs
along three concrete axes: it covers nine \emph{sub-dimensional}
criteria rather than a single overall label, it is collected over
four current text-to-image generators rather than a UI-only context,
and the data is publicly released (DesignPref's data has not been
released publicly at submission time).  TASTE is therefore
complementary to DesignPref rather than substitutive: 
a researcher interested in the broad design domain, with sub-dimensional generation preference, would use TASTE, and one interested in single-label personalised UI preference would use
DesignPref once it releases.

\paragraph{Photo-realistic T2I preference data.}
HPSv2 and HPSv3~\cite{Wu2023HPSv2, Ma2025HPSv3} use pairwise preference
on photo-realistic generation, with one annotator per pairwise comparison
in the training split and a small multi-rater test
split.  Pick-a-Pic~\cite{Kirstain2023PickaPic} collects pairwise
preferences from real users in the
wild.  ImageReward~\cite{Xu2023ImageReward} extends to per-prompt
$k$-way rankings with $k$ between $4$ and $9$ but does not preserve
per-rater identity in the public release.
GenAI-Bench~\cite{Lin2024GenAIBench} provides Likert ratings from
three annotators per item across six T2I
models.  ImagenWorld~\cite{Sun2026ImagenWorld} stress-tests image
generation with explainable human evaluation on real-world tasks.

\paragraph{Multi-dimensional T2I preference.}
RichHF-18K~\cite{Liang2024RichHF} adds per-image scores for
plausibility, alignment, and aesthetics on Pick-a-Pic
outputs.  MHP/MPS~\cite{Zhang2024MPS} collect approximately 918k
pairwise judgments on four orthogonal
dimensions.  VisionReward~\cite{Xu2026VisionReward} extends this to a
binary checklist of around 63 sub-dimensional questions.  In an
adjacent modality, VBench~\cite{Huang2024VBench} and
VBench-2.0~\cite{Huang2025VBench2} establish multi-axis decomposition
as the standard evaluation pattern for video generation, and motivate
the same pattern in still imagery.  All five target photo-realistic
or generic content rather than design.  E-commerce poster evaluation
is closer to design but evaluates AI judges rather than collecting
designer preferences~\cite{Ecomiq2026}.

\paragraph{AIGC quality assessment.}
AGIQA-3K~\cite{Li2023AGIQA3K}, AIGIQA 20K~\cite{Li2024AIGIQA20K}, and AGIN~\cite{Chen2024AGIN} collect Mean Opinion Scores on AI-generated images.  
These target perceptual quality and naturalness rather than design preference, and their public releases either aggregate per-image scores or use a different statistical shape, preventing direct overlay on our histograms.

\paragraph{Classical preference data and statistical machinery.}
Sushi~\cite{Kamishima2003Sushi} and
MovieLens~\cite{Harper2015MovieLens} are canonical multi-rater
preference datasets in the food and movie domains.  Both can be
subsampled to a $(p=4, R=5)$ shape and bracket subjective taste-based
agreement.  Our statistical pipeline builds on Mallows permutation
models~\cite{Mallows1957}, the Bradley--Terry~\cite{BradleyTerry1952}
and Plackett--Luce~\cite{Plackett1975, Luce1959} parametric
formulations, Kendall's $\tau$~\cite{Kendall1938}, and Krippendorff's
$\alpha$~\cite{Krippendorff2004alpha, Krippendorff2018}.

\section{The TASTE dataset}
\label{sec:dataset}

\subsection{Generators, designers, prompts}
We evaluate four current text-to-image models:
FLUX.2 max~\cite{BlackForest2024Flux},
GPT Image 1.5~\cite{OpenAI2024DALLE},
Nano Banana 2~\cite{Google2024Imagen}, and
Seedream 5.0 Lite~\cite{Bytedance2024Seedream}.  Generators are
presented to evaluators behind blind code-names.  The mapping from
code-name to model is recorded but not shown to the rater, which
prevents brand-anchoring effects.

We recruited two disjoint cohorts of professional designers, an
Aesthetics cohort of 5 designers and a Descriptions cohort of 5
different designers.  We refer to evaluators within each cohort as
Designer~A through Designer~E to preserve anonymity in the analyses
that follow.  Each evaluator within a cohort sees every prompt batch
in that cohort, so per-rater identity carries across the cohort's
sub-dimensions.  We paid designers hourly market rates and did not
disclose the statistical analyses applied to their judgments.

\subsection{Sub-dimensional decomposition}
Nine ranking sub-dimensions are collected, organised into the two
cohorts (Table~\ref{tab:dim_layout}).  The criteria were not chosen
by the authors but surfaced from pilot interviews with the same pool
of professional designers who later annotated the data; the exact
interview format is documented in \S\ref{sec:dataset:curation}.  The
Aesthetics cohort covers an overall preference (intent-grounded) and four
sub-criteria the designers consistently flagged as separate cognitive
processes (mood and tone, visual hierarchy, color harmony,
typography).  The Descriptions cohort covers overall preference and
three description-fidelity sub-criteria that designers identified as
verifiable against the prompt (color accuracy, spatial accuracy,
typography).  The two typography criteria are distinct: Aesthetics
Typography rates the typographic craft of the rendered design (font
selection, spacing, sizing hierarchy, alignment), while Descriptions
Typography rates whether the text demanded by the prompt is rendered
correctly (presence, spelling, fit to the brief's directions).  Each sub-dimension was rated on a separate, disjoint
set of 80 prompts.  This design isolates per-dimension signal at the
cost of preventing within-rater within-prompt cross-dimensional
comparison.

For each prompt the rater performs the $\binom{4}{2} = 6$ pairwise
comparisons of the four generator outputs, presented blind side by
side.  The 6 pairwise judgments per (prompt, evaluator) tuple are
then aggregated into a strict 4-way ranking by the collection
pipeline; the exact aggregation procedure is documented in
\S\ref{sec:dataset:curation}.  Each (prompt, evaluator) cell of the
released data is therefore both a set of 6 pairwise comparisons and
a derived 4-way rank, and the analyses in
\S\ref{sec:res} use the derived rank.

\subsection{Hallucination flags}
\label{sec:dataset:halluc}

For the two overall-preference cohorts (Aesthetic Preference and Descriptions Preference, $160$ prompts in total), each evaluator additionally flags every generator's image as \textit{No}, \textit{Minor}, or \textit{Major} hallucination, encoded as $\{0, 1, 2\}$. This yields $1{,}600$ ordinal flags per cohort, distributed across three classes rather than a low-prevalence binary: roughly $55\%$ \textit{No}, $35\%$ \textit{Minor}, and $10\%$ \textit{Major} in both cohorts. Inter-rater agreement is modest. Krippendorff's $\alpha$ with an ordinal distance metric reaches $0.141$ on Aesthetics and $0.165$ on Descriptions, slightly below the $\alpha \approx 0.19$ measured on the same cohorts' rankings. The disagreement is concentrated rather than diffuse: on $57\%$ of Aesthetics items and $52\%$ of Descriptions items, raters split only between \textit{No} and \textit{Minor}, while only $5\%$ split solely between \textit{Minor} and \textit{Major}. Per-evaluator \textit{Major} rates span $3\%$ to $26\%$, reflecting per-rater calibration drift rather than disagreement on which items count as severe failures. We use these flags only as a documented dataset feature in this paper and do not filter the ranking data with them. Tightening the \textit{No}--\textit{Minor} rubric, possibly through automated VQA-style probes in the spirit of recent T2I hallucination evaluation work~\cite{Lim2025IHallA, Liu2025HallucinationSurvey}, is a natural follow-on for the next collection round.

\subsection{Curation details}
\label{sec:dataset:curation}
We hereby describe the data curation and annotation protocol.

\paragraph{Recruitment platform and screening.} 
Designers were recruited through Contra, a professional network of 1.5M creative professionals. Participants comprised experienced graphic designers spanning brand identity, visual systems, web and product design, illustration, and marketing design, with substantial commercial portfolios including work for established technology companies, global brands, and high-growth startups. Designers were eligible for participation based on previous engagement with Contra Labs research studies or by passing the Contra Labs Network eligibility screener, a component of the Contra Labs application completed by all new members. From this pool, selected participants had demonstrated professional graphic design experience through their Contra profiles and prior client work.

\paragraph{Designer demographics.} The 10 participants in this study were located in the US, Portugal, Italy, France, Spain, Pakistan, Armenia, and Jordan. The primary working language of all participants is English. Self-reported specialties, aside from graphic design, include brand design, art direction, advertising design, and AI visual art, with years of experience ranging from 4--10 years. 

\paragraph{Pilot interviews.}
A series of small-scale pilot studies was conducted to refine the questionnaire, study parameters, and evaluated sub-dimensions. The final cohort structure was developed collaboratively between LICA and Contra Labs, informed by prior findings from the Contra Labs Human Creativity Benchmark (HCB) research and successive pilot evaluations. Once the methodology was finalized, the study was conducted at scale.

\paragraph{Prompt sourcing.} All prompts represented in this study were tied to real marketplace outcomes, tied to common deliverables for the creative professional community on Contra. This study focused on prompts from the following subcategories, all sourced from the LICA Dataset \cite{Hirsch2026LICA}: Presentations, flyers, instagram posts, social media, print products, cards \& invitations, posters, art \& design, and logo. A single set of prompts were randomly selected from these categories, proportional to their representation in the LICA dataset.

Each prompt in the LICA dataset has three portions. First, User Intent, a description of the purpose and audience of the output. The second portion was Description, a literal description of the layout and elements of the output. The third portion was Aesthetics, a description of the stylistic qualities and elements of the output. The User Intent and Description portions were taken to make one prompt set for the Description cohort. The same User Intent and corresponding Aesthetics portion was taken to make the second prompt set for the Aesthetics cohort. The participants were randomly divided across these two categories, with 5 designers participating per run.

\paragraph{Image-generation parameters.} The model versions used for these evaluations were: Nano Banana 2 (released February 26, 2026), Flux.2 [max] (released December 16, 2025), GPT Image 1.5 (released December 16, 2025), and Seedream 5.0 Lite (released February 13, 2026). All outputs were generated within the Contra evaluation environment using \href{http://fal.ai/}{fal.ai} and checked to ensure the generation did not fail and the aspect ratio was consistent before presenting to designers.
\paragraph{Annotation interface.} The pairwise comparison was conducted in Contra's evaluation environment. For each set of one prompt and the corresponding four outputs, participants were shown the prompt first with a twenty-second countdown to keep them on the prompt screen. After twenty seconds, or when participants finished reading, they were able to click ``I'm Ready'', which moved them to the pairwise comparison screen. On this screen, they saw the prompt and two images at a time, which they were able to click and zoom into if needed. For selection, they clicked one of two options, which read ``I prefer left'' or ``I prefer right'', and then the next pair of images in the set was presented. They continued this selection until their ranking was settled. There was no timer associated with the selection, but participants were advised to take approximately 2 minutes per evaluation, after reading the prompt. The opening context paragraph shown to designers at the start of every session is reproduced verbatim in Appendix~\ref{app:guidelines}.
\paragraph{Pairwise-to-rank aggregation.} For each (prompt, evaluator) tuple, we collapsed the 6 binary pairwise judgments into a strict 4-way ranking by fitting a Bradley-Terry model and ordering responses by descending estimated strength. We observed no intransitive triples in the released data.
\paragraph{Hallucination rubric.} Designers were advised to complete the hallucination flag with major hallucinations counting as anything outside the scope of the prompt request and unrelated to the prompt. Minor hallucinations counted as elements that were changed or different from intention but still related, and no hallucinations meant that all elements in the design were logically derived from the prompt.
\paragraph{Compensation, consent, and ethics.} Designers were paid a flat project fee that averaged approximately \$90 per hour. Designers in the Description cohort completed 4 evaluations, totalling about 13 hours each (task familiarization and evaluation time) spread over a week-long period. Designers in the Aesthetics cohort completed 5 evaluations, totalling 16 hours each over the same period. All designers gave informed consent through a project contract that documented the use of the collected data for research and product purposes, including external release. The study was conducted as paid market research under Contra Labs' standard research consent procedure and does not fall under the scope of an academic IRB; designers were free to withdraw at any point without penalty. 

\begin{table}[t]
\small
\centering
\begin{tabular}{lllr}
\toprule
Cohort & Sub-dimension & Designers & \# prompts \\
\midrule
Aesthetics    & UI+Ad Preference (holistic) & 5 (cohort A) & 80 \\
Aesthetics    & Mood and Tone Match         & 5 (cohort A) & 80 \\
Aesthetics    & Visual Hierarchy            & 5 (cohort A) & 80 \\
Aesthetics    & Color Harmony              & 5 (cohort A) & 80 \\
Aesthetics    & Typography                  & 5 (cohort A) & 80 \\
\midrule
Descriptions  & Preference (holistic)       & 5 (cohort B) & 80 \\
Descriptions  & Color Accuracy             & 5 (cohort B) & 80 \\
Descriptions  & Spatial Accuracy            & 5 (cohort B) & 80 \\
Descriptions  & Typography                  & 5 (cohort B) & 80 \\
\bottomrule
\end{tabular}
\caption{Cohort structure and per-sub-dimension prompt counts.  The
two cohorts use disjoint pools of professional designers and disjoint
80-prompt batches per sub-dimension.  Each (prompt, evaluator) cell
is a strict 4-way ranking over the four generator outputs, giving
$80 \times 5 \times 4 = 1{,}600$ ratings per sub-dimension.}
\label{tab:dim_layout}
\end{table}

\begin{figure}[t]
\centering
\includegraphics[width=\linewidth]{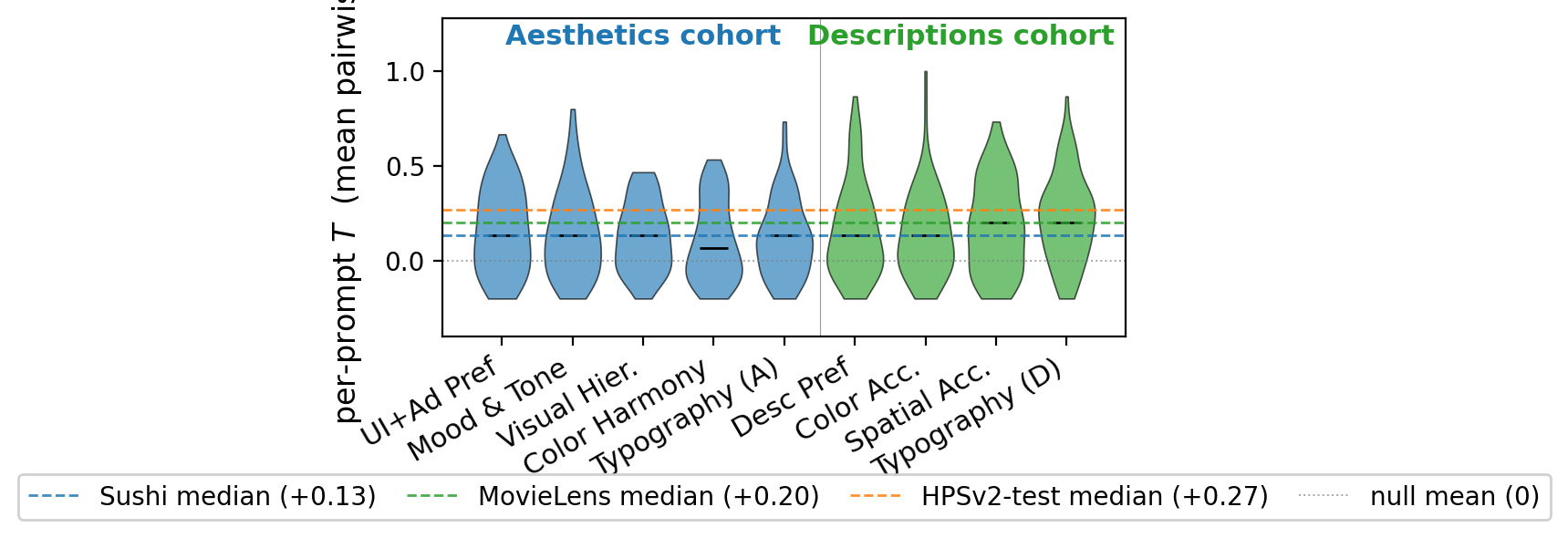}
\caption{Per-prompt $T$ distribution (the mean pairwise Kendall $\tau$
across the 10 evaluator pairs) for each TASTE sub-dimension, with
horizontal lines showing the median $T$ of the three cross-domain
reference anchors.  Aesthetics sub-dimensions are shown in blue and
Descriptions in green.  All nine sub-dimensions sit close to the
Sushi (food) and MovieLens (movies) medians, well below the median of
HPSv2-test restricted to its top-4 generative models, which places
design preference in the subjective-taste regime rather than the
image-quality regime.}
\label{fig:cross_domain}
\end{figure}

\section{Testing for preference signal}
\label{sec:stats}

Before treating TASTE as training data for a preference model, we
test whether the ratings carry meaningful preference signal.  A
dataset of random rankings has the same shape as TASTE and would
survive any conventional descriptive report; the relevant test is
whether the observed agreement exceeds what independent uniform
raters would produce.  We
adopt three complementary signal tests, each addressing a different
mode of disagreement.  Formal definitions, support sets, and null
PMFs are tabulated in Appendix~\ref{app:stats}; the body uses each
statistic intuitively.

\textbf{Kendall's $\tau$ summarises rank correlation between rater
pairs.}  For a single prompt with $R=5$ raters we compute one
Kendall $\tau$ per rater pair and report the per-prompt mean
$T = \tfrac{1}{\binom{R}{2}} \sum_{r<s} \tau(\pi_r, \pi_s)$.
$T = +1$ if all raters agree exactly, $T = 0$ under random ranking,
and $T < 0$ if raters systematically disagree.  At $(p=4, R=5)$ the
support of $T$ is finite and its iid-uniform null PMF has a
closed form, which lets us test the \emph{full histogram} of
per-prompt $T$ values against the null rather than only its mean.

\textbf{Majority-vote probability $\pmax$ asks how often a clear
winner emerges between two items.}  For each pair of generator outputs
$(a, b)$ we count how many of the $R=5$ raters place $a$ above $b$,
call this count $k$, and report $\pmax = \max(k/R, 1-k/R)$.  At $R=5$
this takes values in $\{3/5, 4/5, 5/5\}$.  Larger $\pmax$ means a
sharper majority on a given pair.  Under iid-uniform raters, $\pmax$
follows the symmetric-folded $\mathrm{Bin}(5, 1/2)$ distribution.

\textbf{The Condorcet cycle indicator distinguishes shared-ordering
disagreement from factional disagreement.}  For each prompt we ask
whether the majority pairwise preferences across the four generator
outputs contain an intransitive triple ($A \succ B$, $B \succ C$,
$C \succ A$).  A single weak shared preference order with
rater-specific noise on top suppresses cycles below the iid rate; a
mixture of factions with internally consistent but mutually
orthogonal preferences elevates cycles above the iid rate.  At
$(p=4, R=5)$ the iid-uniform cycle rate is $0.211$ by Monte Carlo,
and we test observed cycle counts against this rate with a binomial
test.

\textbf{Why the three statistics together.}  Kendall's $\tau$ alone
does not separate shared-ordering noise from factional disagreement,
because both can produce similar mean $\tau$.  $\pmax$ alone cannot
diagnose triadic intransitivity.  The cycle indicator alone gives no
information about graded agreement.  Together the three test the
same null (iid-uniform raters) along three independent axes;
rejection on any one is evidence of preference signal, and the
\emph{pattern} of rejection across the three discriminates between
the three competing stories about what causes the observed
disagreement (\S\ref{sec:intro}).  We argue this signal-check
procedure should be standard practice for any new preference dataset
before training begins.

\section{Cross-domain reference anchors}
\label{sec:anchors}

To position TASTE against existing preference data we construct
three reference anchors at the same $(p=4, R=5)$ shape.  Each anchor
is a larger ranking dataset, and for each synthetic prompt we draw
$4$ items and $5$ raters at random and re-rank to $\{1,2,3,4\}$ per
rater on the selected subset.  The resulting $5 \times 4$ rank matrix
is processed by the same pipeline.  Pooling many such
synthetic prompts gives a reference distribution at the same support
as TASTE.

Sushi~\cite{Kamishima2003Sushi} contains preference rankings of $10$
sushi items by approximately 5{,}000 users.  We draw 50 bootstrap
iterations of $10$ panels with $40$ random 4-item subsets per panel,
totalling 20{,}000 samples.

MovieLens~\cite{Harper2015MovieLens} contains $1$-to-$5$ ratings
of films by tens of thousands of users.  We restrict to the top-100
most-rated movies, draw users with sufficient overlap, and emit 50
bootstrap iterations of $12 \times 40 = 480$ samples each.

The HPSv2 test split~\cite{Wu2023HPSv2} contains 400 prompts with
10 fixed annotators per prompt and 9 generative-model images per
prompt, plus a COCO real image on 100 of the prompts.  To make a
fair generated-only comparison at $(p=4, R=5)$ we exclude the COCO
real image and identify the top-4 generative models by mean
aggregated rank across the 400 prompts.  Indices $[1, 2, 6, 8]$ form
a clean cluster with mean ranks of $1.59$ to $2.59$, separated by a
gap of more than two ranks from the next-best (index 0, mean rank
$5.04$).  We hold these four model indices fixed across all prompts
and bootstrap 50 random 5-of-10 rater subsets, giving $50 \times 400
= 20{,}000$ samples at the matching shape.

\begin{table}[t]
\small
\centering
\begin{tabular}{lrrrr}
\toprule
Anchor ($n_{\text{total}}$) & median $T$ & mean pair-$\tau$ & mean $\pmax$ & cycle rate \\
\midrule
Sushi (20{,}000)               & $+0.133$ & $+0.144$ & $0.739$ & $0.107$ \\
MovieLens (24{,}000)           & $+0.200$ & $+0.216$ & $0.764$ & $0.146$ \\
HPSv2-test top-4 (20{,}000)    & $+0.267$ & $+0.302$ & $0.790$ & $0.060$ \\
\midrule
\textit{iid-uniform null}      & $0$      & $0$      & $0.625$ & $0.211$ \\
\bottomrule
\end{tabular}
\caption{Reference anchor descriptive statistics at $(p=4, R=5)$.
All three anchors lie above the iid-uniform null on every statistic.
HPSv2-test (image quality) shows the strongest agreement signal
once weak generators are excluded, while Sushi and MovieLens bracket
the subjective-taste regime.}
\label{tab:anchors}
\end{table}

\section{Results}
\label{sec:res}

\subsection{Every criterion carries preference signal}
\label{sec:res:signal}
The signal check from \S\ref{sec:stats} asks the prior question of
whether TASTE carries preference signal.  Chi-squared
GOF tests on the per-prompt $T$ histogram and the $\pmax$ histogram
against the exact iid-uniform null reject in every TASTE criterion
at $p < 10^{-10}$ on at least one of the two statistics, and in
most criteria on both.  These rejections survive any reasonable
multiple-comparison correction (Bonferroni across $9 \times 2 = 18$
tests requires $p < 0.0028$, comfortably exceeded by the observed
magnitudes).  The cycle indicator adds a
third axis on which agreement is detectable; we treat it separately
in \S\ref{sec:res:cycles} because it discriminates between two
\emph{kinds} of structured agreement rather than between agreement
and noise.  Per-criterion summary statistics appear in
Table~\ref{tab:per_dim}.

\begin{table}[t]
\small
\centering
\begin{tabular}{llrrrr}
\toprule
Cohort & Sub-dim & median $T$ & mean pair-$\tau$ & mean $\pmax$ & cycle rate \\
\midrule
Aesthetics   & UI+Ad Preference   & $+0.133$ & $+0.159$ & $0.744$ & $0.150$ \\
Aesthetics   & Mood and Tone      & $+0.133$ & $+0.147$ & $0.737$ & $0.125$ \\
Aesthetics   & Visual Hierarchy   & $+0.133$ & $+0.128$ & $0.734$ & $0.087$ \\
Aesthetics   & Color Harmony     & $+0.067$ & $+0.103$ & $0.723$ & $0.138$ \\
Aesthetics   & Typography         & $+0.133$ & $+0.119$ & $0.733$ & $0.075$ \\
\midrule
Descriptions & Preference         & $+0.133$ & $+0.163$ & $0.745$ & $0.150$ \\
Descriptions & Color Accuracy    & $+0.133$ & $+0.144$ & $0.741$ & $0.113$ \\
Descriptions & Spatial Accuracy   & $+0.200$ & $+0.182$ & $0.750$ & $0.150$ \\
Descriptions & Typography         & $+0.200$ & $+0.224$ & $0.767$ & $0.062$ \\
\bottomrule
\end{tabular}
\caption{Per-sub-dimension descriptive statistics ($n = 80$ prompts each).
Every dimension rejects the iid-uniform null on $T$, pair-$\tau$, and
$\pmax$ at $p < 10^{-10}$.  Cycle-rate tests vary by dimension.
Description-fidelity sub-dimensions (Spatial Accuracy, Typography)
show the strongest agreement signal.}
\label{tab:per_dim}
\end{table}

\subsection{Where designer agreement sits across domains}
\label{sec:res:cross}
The first signal-test statistic, Kendall's $\tau$ aggregated as
per-prompt $T$, places TASTE against the three reference datasets.
Figure~\ref{fig:cross_domain} overlays the per-prompt $T$
distribution of every TASTE criterion on the median $T$ of the
three anchors.  All nine criteria have median $T$ between $+0.067$
(Color Harmony) and $+0.200$ (Spatial Accuracy and Descriptions
Typography), and the band $[+0.067, +0.200]$ sits inside the
Sushi--MovieLens band $[+0.133, +0.200]$.  HPSv2-test on its top-4
generators sits at median $T = +0.267$, above even TASTE's
strongest criterion.  Designers ranking graphic design therefore
agree at roughly the rate users agree on movies, and below the
rate crowdworkers agree when ranking photo-style image quality.  A
reward model trained on photo-style preferences is unlikely to fit
TASTE zero-shot, and soft labels or per-criterion fits are needed.

\subsection{Cycle rates rule out factional disagreement}
\label{sec:res:cycles}
\begin{figure}[t]
\centering
\includegraphics[width=\linewidth]{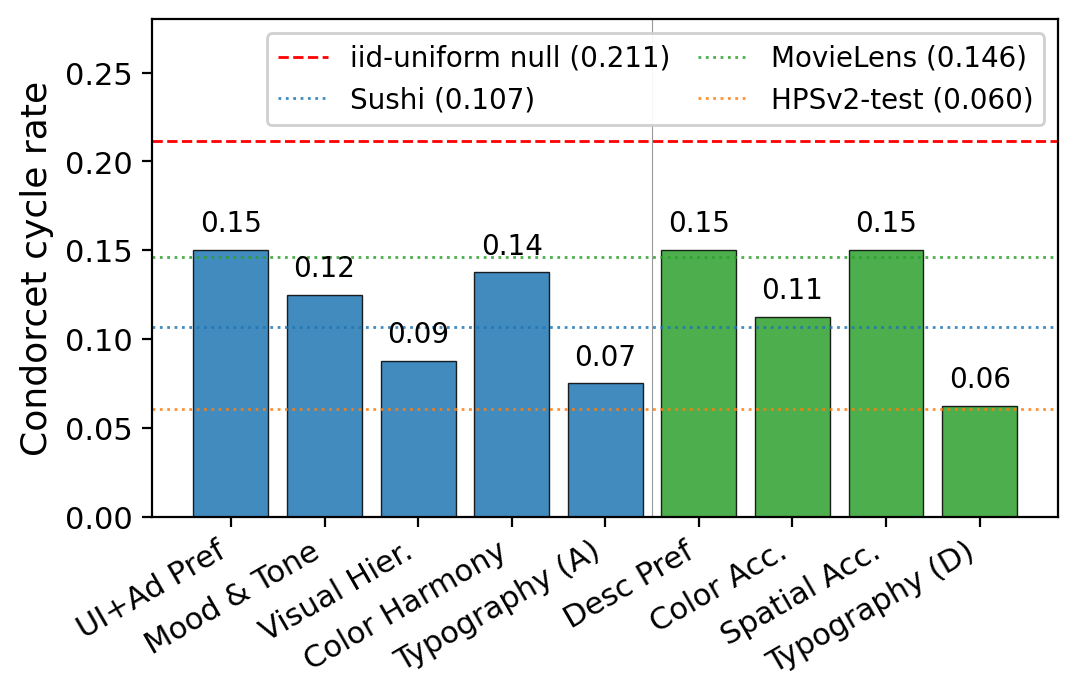}
\caption{Per-sub-dimension Condorcet cycle rate.  All nine
sub-dimensions sit at or below the iid-uniform null rate (red
dashed at $0.211$), the qualitative signature of a Mallows weak
common ordering rather than a mixture-of-orderings.  Horizontal
dotted lines show reference cycle rates for Sushi, MovieLens, and
HPSv2-test top-4.}
\label{fig:cycle_rates}
\end{figure}

The third signal-test statistic, the Condorcet cycle indicator,
discriminates between two structured-agreement stories that
$\tau$ and $\pmax$ alone cannot tell apart.  A factional model
predicts cycle rates above the iid-uniform rate; a weak shared
ordering predicts cycle rates below it.  For each criterion we
test the observed cycle count against the binomial null
$\mathrm{Bin}(80, 0.211)$ with a two-sided test.  All nine observed
rates lie at or below the null rate of $0.211$
(Figure~\ref{fig:cycle_rates}).  Four criteria reject the null at
$\alpha = 0.05$, in each case with the rate significantly
\emph{below} it (Aesthetics Visual Hierarchy at $p = 0.006$,
Aesthetics Typography at $p = 0.001$, Descriptions Color Accuracy
at $p = 0.028$, Descriptions Typography at
$p = 4.9 \times 10^{-4}$).  The remaining five criteria do not
reject the null individually at $n = 80$, but the joint pattern
across all nine is strongly one-sided; no criterion shows the
elevated-cycle signature that a factional model would generate.
Combined with the left-shifted $T$ histogram from
\S\ref{sec:res:cross}, this rules out the random-rater and
factional hypotheses for every criterion.  A weak shared preference
order with rater-specific noise is the residual explanation.  An
aggregate model with agreement-weighted soft labels (such as
Bradley--Terry or an HPS-style cosine model) should therefore work
better than a mixture-of-experts setup on TASTE, since the mixture
has no factional structure to recover.

\subsection{Signal level varies markedly across criteria}
\label{sec:res:subdim}
\begin{figure}[t]
\centering
\includegraphics[width=\linewidth]{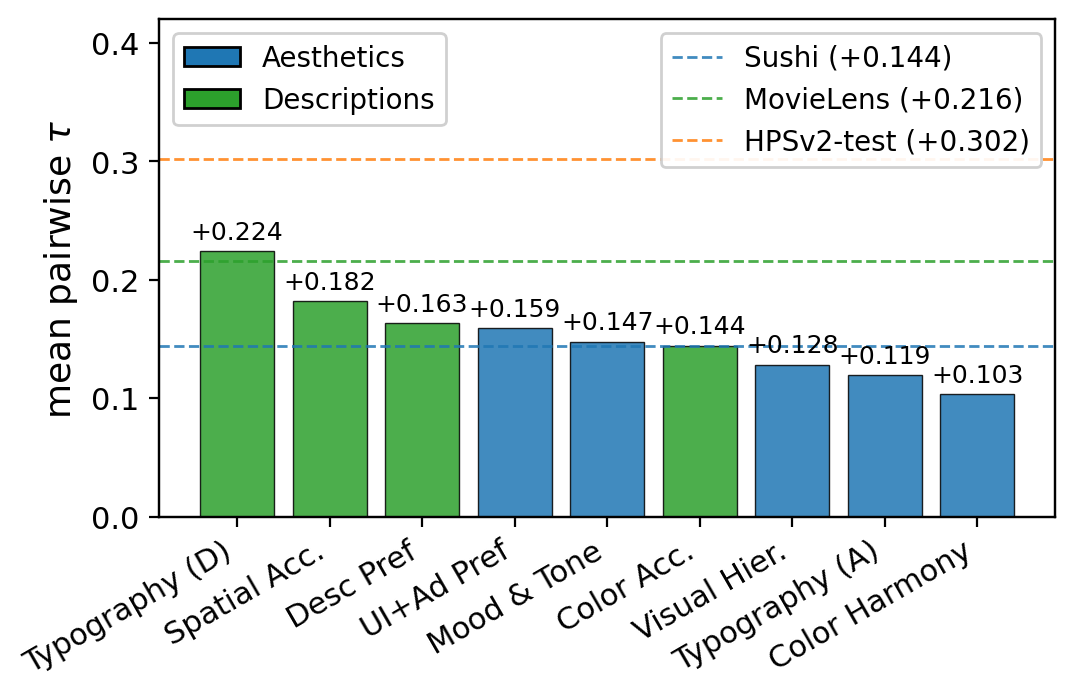}
\caption{Per-sub-dimension mean pairwise $\tau$, sorted descending,
color-coded by cohort.  Description-fidelity sub-dimensions (green)
dominate the top of the ordering; aesthetic sub-dimensions (blue)
dominate the bottom.  Reference anchor $\tau$'s are shown as
dashed lines.}
\label{fig:subdim}
\end{figure}

The signal level itself, measured by mean pairwise $\tau$, varies
within TASTE by a factor of roughly $2.2$ across criteria
(Figure~\ref{fig:subdim}).  Aesthetics Color Harmony sits lowest at
$+0.103$; Descriptions Typography sits highest at $+0.224$.  The
four description-fidelity criteria (Descriptions Typography, Spatial
Accuracy, Preference, Color Accuracy) all sit at or above the
Sushi anchor; the five aesthetic criteria cluster below it, with
Color Harmony at the bottom.  This ordering is consistent with
the observation that typography rules and spatial layout
constraints are more crisply codified than color harmony or mood,
giving them sharper inter-rater agreement.  For a downstream
modeller, description-fidelity criteria are the natural starting
point for criterion specialists, and aesthetic criteria are better
treated as soft preferences.

\subsection{Per-evaluator agreement structure}
\label{sec:res:outliers}
\begin{figure*}[t]
\centering
\begin{tabular}{cc}
\includegraphics[width=0.46\linewidth]{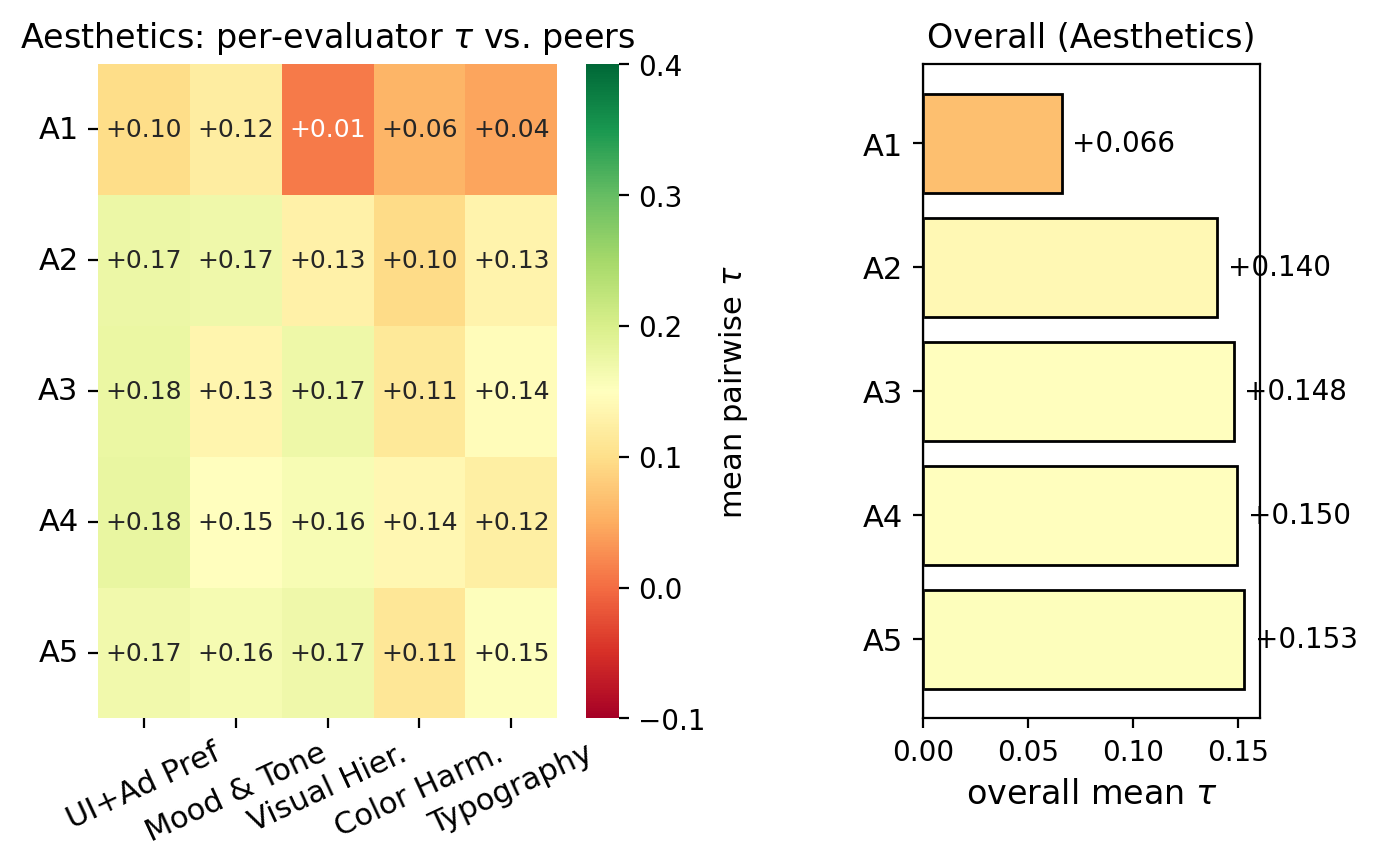} &
\includegraphics[width=0.46\linewidth]{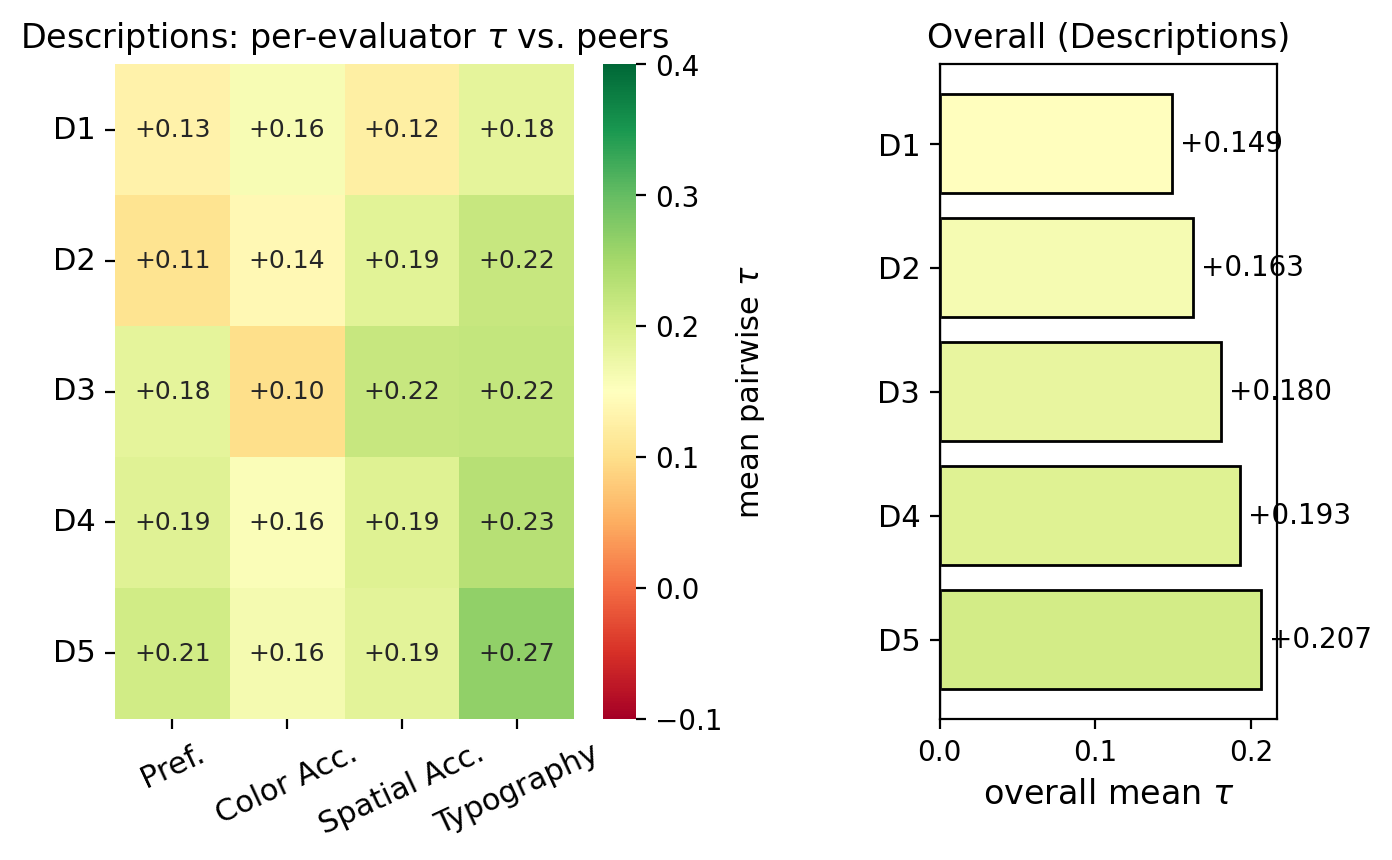} \\
\end{tabular}
\caption{Per-evaluator agreeableness on the Aesthetics (left) and
Descriptions (right) cohorts.  Each cell of the heatmap reports the
mean pairwise $\tau$ between an evaluator and the other four in the
same cohort, averaged over the 80 prompts of one sub-dimension.  Bar
plots show the per-evaluator overall mean.  Within each cohort
evaluators are ordered by ascending overall mean $\tau$ and re-labelled
Designer~A through Designer~E.  Designer~A is the consistent outlier
in both cohorts; Designer~E is the most agreeable.  The original
designer identities are anonymised for double-blind review.}
\label{fig:eval_outliers}
\end{figure*}

We compute the $5 \times 5$ per-pair mean $\tau$ matrix for each
sub-dimension and identify the most-disagreeing pair per dimension.
In the Aesthetics cohort, one designer (the least-agreeing
designer, re-labelled Designer~A in
Figure~\ref{fig:eval_outliers}) participates in the most-disagreeing
pair for all five sub-dimensions and never in the most-agreeing
pair.  The Descriptions cohort shows the same pattern for the
analogous designer, who participates in the most-disagreeing pair
for three of four sub-dimensions.  This pattern is not an artefact
of a single dimension and emerges robustly across the sub-dimensions
of each cohort, identifying these designers as candidates for
agreement-weighted training or a second annotation pass rather than
exclusion.  The finding is consistent with the cycle-rate result:
a single per-cohort low-agreement rater raises the disagreement
floor of every dimension uniformly without producing the structured
factional disagreement that would push cycle rates above the null.

\subsection{Pre-trained preference models and VLM judges}
\label{sec:res:vlmjudge}

We benchmark two classes of pre-trained models on TASTE.
The first class is dedicated preference and aesthetic scorers
trained on AI-image preferences: HPSv2.1~\cite{Wu2023HPSv2},
PickScore-v1~\cite{Kirstain2023PickaPic}, and
LAION-Aesthetic-V2.  The second class is vision-language models
prompted with the brief and both candidate images to pick the
better one, the dominant evaluation mode in recent text-to-image
judging benchmarks.  Our open-weight VLM slate matches the
closest comparable benchmark~\cite{Goyal2026DistortBench} and
covers Qwen3-VL-Instruct at $4$B, $8$B, and $32$B parameters,
Gemma-3-27B-it, Kimi-VL-A3B-Instruct, and InternVL3.5-14B, six
models across four families with a $10\times$ active-parameter
range and both dense and MoE architectures.

The metric is agreement with the 5-designer majority on the
$\binom{4}{2}{=}6$ ranking pairs per prompt, with human-tie pairs
excluded from the denominator, so chance is $0.5$.  The VLM
judges follow the field-standard pairwise protocol with three
components.  Inline \texttt{Image A:} and \texttt{Image B:} labels
are placed immediately before each
image~\cite{Tian2025MultiImageBias}.  The model is asked for a
structured JSON answer~\cite{Goyal2026DistortBench}.  Position
bias is corrected via MT-Bench S1~\cite{Zheng2023MTBench}, in
which each pair is judged in both image orders and
order-inconsistent verdicts are counted as ties at $0.5$.  Each
VLM produces $8{,}640$ inferences (4{,}320 pairs $\times$ two
orders).  Methodological details, including eight question
paraphrases per criterion and the exact prompt template, are in
Appendix~\ref{app:vlmjudge}.  The dedicated scorers are
deterministic given (image, brief), so order correction does not
apply.

\begin{table}[t]
\centering
\small
\setlength{\tabcolsep}{6pt}
\begin{tabular}{llrr}
\toprule
Model & Family & Active & Macro acc. \\
\midrule
\multicolumn{4}{l}{\textit{Open-weight VLM judges (six models)}}\\
Qwen3-VL-4B-Instruct    & Qwen            & 4\,B  & 0.530 \\
Qwen3-VL-8B-Instruct    & Qwen            & 8\,B  & 0.539 \\
Qwen3-VL-32B-Instruct   & Qwen            & 33\,B & 0.536 \\
Gemma-3-27B-it          & Google          & 27\,B & 0.528 \\
Kimi-VL-A3B-Instruct    & Moonshot (MoE)  & 3\,B  & 0.509 \\
InternVL3.5-14B         & OpenGVLab       & 14\,B & 0.525 \\
\midrule
\multicolumn{4}{l}{\textit{Dedicated preference / aesthetic scorers}}\\
HPSv2.1                 & reward          &       & 0.543 \\
PickScore-v1            & reward          &       & 0.522 \\
LAION-Aesthetic-V2      & aesthetic head  &       & 0.499 \\
\bottomrule
\end{tabular}
\caption{Agreement with the 5-designer majority winner, averaged
over the nine TASTE criteria ($n{=}4{,}320$ pairs each).
Chance is $0.5$.}
\label{tab:vlm-baselines}
\end{table}

\begin{table*}[t]
\centering
\footnotesize
\setlength{\tabcolsep}{4pt}
\begin{tabular}{lrrrrrrrrr}
\toprule
Criterion & Q-4B & Q-8B & Q-32B & G-27B & K-3B & I-14B & HPSv2.1 & PickScore & LAION \\
\midrule
aesthetics\_color\_harmony & 0.506 & 0.523 & 0.531 & 0.525 & 0.502 & 0.527 & 0.571 & 0.566 & 0.545 \\
aesthetics\_mood           & 0.525 & 0.554 & 0.557 & 0.550 & 0.523 & 0.541 & 0.590 & 0.583 & 0.564 \\
aesthetics\_preference     & 0.575 & 0.564 & 0.593 & 0.578 & 0.535 & 0.545 & 0.561 & 0.534 & 0.495 \\
aesthetics\_typography     & 0.517 & 0.504 & 0.518 & 0.513 & 0.497 & 0.519 & 0.504 & 0.566 & 0.526 \\
aesthetics\_visual\_hier   & 0.511 & 0.509 & 0.482 & 0.502 & 0.488 & 0.498 & 0.487 & 0.514 & 0.547 \\
descriptions\_color\_acc   & 0.530 & 0.533 & 0.527 & 0.496 & 0.512 & 0.521 & 0.542 & 0.477 & 0.470 \\
descriptions\_preference   & 0.532 & 0.554 & 0.551 & 0.541 & 0.510 & 0.539 & 0.577 & 0.504 & 0.468 \\
descriptions\_spatial\_acc & 0.544 & 0.536 & 0.528 & 0.509 & 0.502 & 0.514 & 0.523 & 0.446 & 0.468 \\
descriptions\_typography   & 0.532 & 0.571 & 0.540 & 0.540 & 0.506 & 0.524 & 0.531 & 0.508 & 0.411 \\
\bottomrule
\end{tabular}
\caption{Per-criterion agreement with the designer majority for all
nine systems.  Columns: Q $=$ Qwen3-VL, G $=$ Gemma-3,
K $=$ Kimi-VL-A3B, I $=$ InternVL3.5.  $n{=}480$ pairs per criterion.}
\label{tab:vlm-percrit}
\end{table*}

\paragraph{The accuracy ceiling is class-independent.}
Table~\ref{tab:vlm-baselines} shows all nine systems lie within a
4.4-point band, from $0.499$ (LAION-Aesthetic-V2) to $0.543$
(HPSv2.1), at most 4.3 points above chance, and none reaches
$0.55$.  The ceiling holds across model classes, within-family
scaling (Qwen3-VL at three sizes), across families, and across
architectures (dense and MoE).  HPSv2.1, trained on over 640K
human pairwise comparisons of AI-generated images, is the best
system in the table and reaches only $0.543$.  The T2I-preference
signals these models were trained on do not transfer cleanly to
designer judgment on TASTE.  This is the strongest version of the
cross-domain claim from \S\ref{sec:anchors}.

\paragraph{Per-criterion accuracy reveals model-class
specialisation.}  Table~\ref{tab:vlm-percrit} breaks the macro
down by criterion.  LAION-Aesthetic-V2, an image-only scorer with
no text conditioning, falls to $0.411$ on
$\text{descriptions\_typography}$, the only sub-chance value in
the slate, while reaching $0.547$ on
$\text{aesthetics\_visual\_hier}$.  This is the signature of an
aesthetic-only model failing on text-fidelity tasks, which the
macro number hides.  PickScore-v1 shows a milder version of the
same effect.  HPSv2.1 is balanced across cohorts.

\paragraph{Position bias and content perception trade off; net
accuracy is flat.}  For the VLM judges, the position-bias rate is
the fraction of pairs whose verdict is unchanged when image order
is flipped, and it ranges from $0.44$ (Gemma-27B) to $0.87$
(Kimi-VL-A3B) across the slate.  Restricting to the
order-consistent fraction of each model's verdicts, agreement with
the designer majority rises to $0.55{-}0.66$.  Across the six VLMs,
position-bias rate and this conditional accuracy are strongly
positively correlated (Spearman $\rho{=}{+}0.94$, $p{=}0.005$):
models with more conservative answer policies default to position~A
more often \emph{and} show stronger perception when they do commit.
The two effects cancel and pin net agreement into the
$[0.509, 0.539]$ band regardless of scale, family, or
architecture.  Within Qwen, scaling reduces position bias
monotonically ($0.775 \to 0.679 \to 0.463$ at $4$B, $8$B, $32$B),
and at the largest dense scale we test the operating point is
$\approx 0.45$ for both flagship-dense families (Qwen-32B and
Gemma-27B), but the macro accuracy does not improve.  Scaling moves
the slate along a bias-versus-perception frontier without moving
overall accuracy.  Per-criterion position-bias rates and the full
diagnostic table are in Appendix~\ref{app:vlmjudge}.

These results establish a baseline on the dataset.  Designer
preference is not solved by current off-the-shelf systems regardless
of model class, training data scale, or architecture family.  The
bottleneck is calibration rather than perception in the narrow
sense: order-consistent VLM verdicts agree with designers above
$0.6$ on the better models, but positional priors absorb the
remaining accuracy budget, and neither prompt engineering,
within-family scaling, nor cross-family substitution closes this
gap.  These numbers are the off-the-shelf baseline against which
\S\ref{sec:premodel} reports what a small model trained directly
on TASTE achieves.

\section{Preference-model analysis}
\label{sec:premodel}

Given the preference data, we turn to two model-side questions: \emph{what is the right ceiling} for a model trained to imitate the panel, and \emph{how close can a small, modular preference head reach} when trained on that data alone. 
Both depart from standard classification practice for non-trivial reasons.  
Labels are aggregated from a small panel of raters with measurable disagreement, and roughly half of all pairwise comparisons sit in a 3-2 split where even a perfect predictor cannot distinguish a true preference from a coin flip.
Reporting a single val-accuracy number against such labels is uninterpretable without a calibrated upper bound, and is misleading without a per-bucket breakdown.

We separate these two concerns.  
\S\ref{sec:premodel:eval} gives the calibrated ceilings against which any model trained on this data should be reported; \S\ref{sec:premodel:train} describes the modular head we trained, the architectural choice that lifted accuracy beyond a flat regularization sweep, and the four downstream uses the model unlocks.

\subsection{Evaluation diagnostics and the leave-one-out ceiling}
\label{sec:premodel:eval}

\textbf{Ceiling 1: leave-one-out human imitation.}  
A model trained on the preference data is by construction trying to predict the modal vote of the five-rater panel.  
The natural upper bound is therefore the accuracy of a \emph{single} rater against the majority of the other four.  
With $R = 5$ raters this leave-one-out (LOO) quantity for a
pair $p$ with vote counts $(v_a, v_b)$ is
\[
\textsc{loo}_p \;=\; \max(v_a, v_b) \,/\, R,
\]
which is exactly the agreement statistic of \S\ref{sec:stats} and, averaged over pairs, gives the expected accuracy of a randomly chosen rater on the panel's task.  
Across all nine ranking criteria, \(\overline{\textsc{loo}} = 0.741\) (range $[0.723, 0.767]$; D+typography is highest, A+color-harmony lowest).  
A model that hits 0.741 has matched an annotator on average; a model that exceeds 0.741 commits more confidently on the cases where the average annotator would waffle.

\textbf{Ceiling 2: oracle on decisive, chance on ambiguous.}  
A second, criterion-agnostic upper bound assumes oracle behavior on
every pair where a strict 4-1 or 5-0 majority exists and chance
behavior on 3-2 splits.  
With bucket fractions $(f_{\text{unanim}}, f_{\text{majority}}, f_{\text{split}}) = (0.173, 0.361, 0.466)$ this gives
\[
\textsc{cap} \;=\; f_{\text{unanim}} + f_{\text{majority}} + 0.5\,f_{\text{split}} \;=\; 0.767.
\]
The two ceilings are close because LOO and \textsc{cap} measure the same scarce signal from different sides; the gap $\textsc{cap} - \overline{\textsc{loo}} = 0.026$ is the amount by which a disciplined predictor that commits on ambiguous pairs can beat a single-rater imitator.

\textbf{Practical implication.}  
Any val accuracy on that data should be reported alongside 
(i) the bucket-stratified accuracy breakdown, 
(ii) $\overline{\textsc{loo}}$ for the criteria in scope, and 
(iii) \textsc{cap}.  
Absolute accuracy on its own is misleading: on this data, any pairwise-prediction accuracy above 0.78 is suspect (very likely val leakage or a misaligned ground truth), and any number below 0.55 sits at the random-rater floor.

\subsection{Model training, results, and use}
\label{sec:premodel:train}

\paragraph{Architecture.}  
We train a modular preference head on top of a frozen vision-language encoder (Qwen3-VL-Embedding-2B~\cite{qwen3vlembedding}), keeping the encoder out of the optimization budget.  
The architecture is one small MLP per ranking dimension, dispatched at training time by the row's criterion slug, plus one binary head for the hallucination labels.  
All heads consume the same fused $[t,\, i,\, t \odot i,\, |t - i|]$ representation of the $\ell_2$-normalized text ($t$) and image ($i$) embeddings, where $\odot$ is the element-wise product; concatenating the unimodal vectors with explicit alignment ($t \odot i$) and disagreement ($|t-i|$) terms avoids forcing the MLP to learn these cross-modal interactions from scratch.

\paragraph{Training objectives.}  
The per-dimension scoring heads are trained with an agreement-weighted Bradley-Terry (BT) loss~\cite{bradley1952rank}.  
The BT model treats each pairwise comparison as a Bernoulli trial whose success probability is \(\sigma(\tau \cdot (s_a - s_b))\), where \(\sigma\) is the logistic function, \(s_a\) and \(s_b\) are the head's scalar scores for the two images, and \(\tau = \exp(\textsl{logit\_scale})\) is a single learnable temperature.\footnote{
The learnable temperature rescales the potentially small score differences produced by the frozen-backbone scoring head, allowing the Bradley–Terry model to convert relative rankings into calibrated preference probabilities with sharper discrimination.}  
The per-pair loss is the binary cross-entropy of this probability against the empirical win-rate of \(a\) across raters, weighted by the panel's per-pair agreement so unanimous and near-unanimous cases contribute more.
The hallucination head is trained with binary cross-entropy against the per-asset positive-vote share.  
Both losses are minimized jointly with the two batch streams alternated.  
Training uses \emph{per-evaluator rows} (one labeled pair per annotator) rather than aggregated soft labels, which gives the BT loss \(\approx 5\times\) more independent gradient signal per pair without losing the consensus cases.

\paragraph{Architectural choices.}  
A baseline sweep varying head width (\(\in\{64,128,256\}\)), dropout (\(\in\{0.2,0.3\}\)), weight decay (\(\in\{0,0.05,0.1\}\)), and label form (hard vs.\ soft) is essentially flat at val accuracy \(0.583 \pm 0.011\), so none of the standard regularization knobs produces a change above the noise floor.  
This points to a \emph{structural} ceiling rather than a capacity one.  
In the architecture described above, each image is scored independently by a shared head \(f\) before the comparison happens, so the BT logit is forced into the form
\[
\tau \cdot \bigl( f(t, i_a) - f(t, i_b) \bigr).
\]
Any decision rule that requires \emph{joint} attention to features of \(i_a\) and \(i_b\), for example, "\(i_a\) wins because it renders text legibly while \(i_b\) does not, given the prompt asks
for typography", cannot be expressed as a difference of two prompt-conditioned per-image scalars, no matter how wide \(f\) is.
We lift this restriction with a \emph{pairwise-difference} scoring head that computes \(g(t, i_a, i_b)\) directly.  
The head consumes
\[
[t,\, i_a,\, i_b,\, i_a - i_b,\, |i_a - i_b|,\, t \odot (i_a - i_b)],
\]
where the cross-term \(t \odot (i_a - i_b)\) gives explicit access to the linear directions in embedding space along which the prompt rewards a difference between the two images.  
The hallucination head, which operates on a single image, retains the original 4-chunk fusion.

\paragraph{Results.}  
We report two views of accuracy: the overall val accuracy aggregated across all ranking dimensions, and the per-bucket val accuracy that separates the easy 5-0 cases from the inherently ambiguous 3-2 splits.  The pairwise-difference head reaches \(\mathbf{0.611}\) overall, \(+0.028\) absolute over the best frozen-backbone scalar baseline, and the first configuration in our sweep to clear the noise floor.  
Anchored against the diagnostic ceilings of \S\ref{sec:premodel:eval}
(Table~\ref{tab:premodel:results}), the model still sits \(\approx 13\) absolute points below the single-rater LOO ceiling \(\overline{\textsc{loo}} = 0.741\), so substantial headroom remains against the human-imitation target; versus the random-rater baseline of \(0.5\) it has captured roughly half of the available signal.

The per-bucket breakdown locates this signal precisely.  
On unanimous (5-0) pairs the model scores \(0.653\), well below the
trivial LOO ceiling of \(1.0\); on 4-1 majority pairs it scores
\(0.600\) against a per-bucket LOO ceiling of \(0.800\); on 3-2
splits it scores \(0.602\), within noise of the per-bucket LOO
ceiling of \(0.600\).  
The model's gap to the LOO ceiling thus shrinks monotonically as the panel becomes more ambiguous (\(0.35 \rightarrow 0.20 \rightarrow 0.00\)), and on the ambiguous half of the dataset \emph{the head is already as accurate as a randomly chosen human annotator at predicting the panel's modal vote.}  
The scalar baseline, by contrast, sits at \(0.558\) on splits, roughly the midpoint between chance and the LOO ceiling.  
The \(+0.028\) overall gain is therefore driven almost entirely by the split bucket (\(0.558 \rightarrow 0.602\)), with a small regression on unanimous (\(0.677 \rightarrow 0.653\)).  
This is the opposite of what we expected (unanimous pairs ought to be easy), and is consistent with the head learning a sharper, more confident decision rule that occasionally overshoots on already-clean cases.
Averaged across the seven ranking dimensions the model's Kendall \(\tau\) against the modal annotator is \(0.15\), broadly consistent with the per-criterion signal-strength.

\begin{table}[t]
\centering
\caption{Preference-model val accuracy by panel agreement bucket.
LOO ceilings follow \(\textsc{loo}_p = \max(v_a, v_b)/R\) with
\(R=5\).  "Best scalar baseline" is the sweep winner
(though all candidates were flat in \([0.576,\,0.583]\) across four scalar configurations); rows 2--4 vary only the head input.  CCP = criterion-conditioned prompt; "pairwise" replaces \(f(t,i_a)-f(t,i_b)\) with the joint head \(g(t,i_a,i_b)\) of \S\ref{sec:premodel:train}.  Bold = best model per column.}
\label{tab:premodel:results}
\small
\begin{tabular}{lcccc}
\toprule
& Overall        & Unanim (5-0)   & Major.\ (4-1)  & Split (3-2)    \\
\midrule
\multicolumn{5}{l}{\emph{Reference points}} \\
Random rater                           & 0.500          & 0.500          & 0.500          & 0.500          \\
Single-rater LOO ceiling               & 0.741          & 1.000          & 0.800          & 0.600          \\
\midrule
\multicolumn{5}{l}{\emph{Frozen-backbone scoring head (ours)}} \\
Best scalar baseline                   & 0.583          & 0.677          & 0.564          & 0.558          \\
\quad + criterion-conditioned prompt   & 0.580          & \textbf{0.685} & 0.584          & 0.529          \\
\quad + pairwise-difference fusion     & \textbf{0.611} & 0.653          & \textbf{0.600} & \textbf{0.602} \\
\quad + CCP and pairwise               & 0.586          & 0.629          & 0.552          & 0.599          \\
\bottomrule
\end{tabular}
\end{table}

The bottom four rows of Table~\ref{tab:premodel:results} hold
capacity and training pipeline fixed and vary only the head
input.  The criterion-conditioned prompt is a wash on its own
(\(-0.003\) val), which indicates that the criterion information
was already accessible to the encoder and the gain from the
pairwise head is not a downstream effect of better criterion
conditioning.  Combining the criterion-conditioned prompt with the
pairwise head regresses to \(0.586\) (vs.\ \(0.611\) for pairwise
alone), so the prompt change actively interferes with the pairwise
mechanism rather than complementing it; the architectural lever
and the prompt lever are not independent and should not be
stacked.

\paragraph{Scope and ablations.}
The pairwise-head gain is obtained with the encoder frozen.  A
LoRA-adaptation experiment on the encoder yielded val accuracy
$0.528$, below the frozen-MLP baseline of $0.583$, so we have no
evidence that fine-tuning the encoder improves preference accuracy
on this data.  Augmenting the text input with criterion-specific
prefixes did not help in a controlled comparison ($< 0.01$ val
change vs.\ baseline) and actively interfered with the pairwise
mechanism when combined; the relevant criterion information is
already accessible to the encoder, and the bottleneck is on the
head side rather than the prompt side.

\paragraph{Use cases.}  
A preference model supports four downstream uses.
(i) \emph{Best-of-\(N\) selection}: given \(N\) candidate generations for a prompt, pick the highest-scoring image under the weighted sum of per-dimension scores most relevant to the deployment (e.g.\ uniform for general use, typography-weighted for poster generation).
(ii) \emph{Reward modeling} for DPO~\cite{Rafailov2023DPO} or RLHF-style fine-tuning of design-oriented T2I models: the per-dimension heads provide a multi-objective reward instead of the single scalar typical of HPSv2 / HPSv3 reward heads~\cite{Wu2023HPSv2, Ma2025HPSv3}, so the optimiser can trade off, e.g.\ typography accuracy against color harmony explicitly.
(iii) \emph{Designer-anchored evaluation} of new T2I models on a held-out data subset: per-dimension Kendall \(\tau\) against the modal annotator gives a quality report that complements task-level benchmarks such as GDB~\cite{Deganutti2026GDB} and generic preference scores by reporting where, axis-wise, the model diverges from designer preference.
(iv) The hallucination head functions as a stand-alone filter for production pipelines, with lower precision and higher recall than human inspection but cheap enough to run on every generation as a first-pass quality gate.

\section{Discussion}
\label{sec:disc}

TASTE records designer preference along nine sub-dimensions of design quality on outputs from four current
text-to-image models, in a form that complements LICA's structural data and Graphic-Design-Bench's task suite.  The three-statistic signal check confirms that the data carries learnable preference signal at every criterion, and the cross-domain anchors locate that signal between subjective taste tasks and photo-style image quality.

The agreement structure also drives concrete modeling choices.
Cycle rates below the iid-uniform null
(\S\ref{sec:res:cycles}) rule out a factional regime, so an
aggregate reward model with agreement-weighted soft labels is the
natural baseline and a mixture-of-experts approach has no factional
structure to recover.  Signal-level heterogeneity across criteria
(\S\ref{sec:res:subdim}) points to specialist heads or per-criterion
loss terms on description-fidelity axes and broad soft preferences
on aesthetic axes.  Per-evaluator outliers
(\S\ref{sec:res:outliers}) admit standard outlier-aware
re-weighting.

\textbf{Signal checks before training.}  The check we apply to
TASTE (Kendall's $\tau$, $\pmax$, cycle indicator, all against
exact $(p=4, R=5)$ nulls and three subsampled reference
distributions) is generic.  It can be applied to any new preference
dataset whose ranking shape can be subsampled to a small $(p, R)$
panel, and answers the prior question of whether the data carries
learnable signal.  We recommend it as a routine prior step
to model training rather than as a TASTE-specific contribution.

\textbf{Limitations.}  Five raters per prompt is the smallest $R$ at
which $\binom{R}{2}=10$ pair-$\tau$ values produce a usable
histogram; larger $R$ would tighten the binomial $\pmax$ null and
lift power on individual prompts.  Each criterion is rated on a
disjoint 80-prompt batch, which forecloses within-rater
within-prompt cross-criterion analysis but bounds the annotation
cost.  Designers come from a single platform and prompts are
sourced in English, so cross-cultural and cross-lingual preference
variation is not captured.  Several criteria that designers care
about in practice are not part of the present collection,
including accessibility (WCAG-style contrast and legibility), brand
consistency, motion and animation (covered structurally by
LICA~\cite{Hirsch2026LICA}), and audience-appropriateness; future
collection rounds should extend the criterion set in those
directions.

\textbf{Release plan.}  The 1{,}600 ratings per criterion (both
the $\binom{4}{2} = 6$ pairwise judgments and the derived 4-way
rankings), the 1{,}600 ordinal hallucination flags per
overall-preference cohort, the analysis pipeline, the per-criterion
distribution-test reports, and the appendix's exact null PMFs will
be released under a permissive licence on a public dataset host
upon acceptance, alongside an evaluation script that runs the
three-statistic signal check on a user-supplied ranking dataset of
arbitrary $(p, R)$ shape.

\section{Conclusion}
\label{sec:conc}

We release TASTE, a designer-annotated multi-dimensional preference dataset for AI-generated graphic design covering nine criteria across two disjoint cohorts of professional designers on outputs from four current text-to-image models. Together with LICA and Graphic-Design-Bench, TASTE adds the missing designer-side preference layer to the graphic-design generation stack. We complement the dataset with a three-statistic signal check (Kendall's $\tau$, majority-vote probability $\pmax$, and Condorcet cycle existence) against exact null distributions at $(p=4, R=5)$, confirming that learnable signal is present in every criterion, placing designer agreement between subjective taste tasks and photo-style image quality, and ruling out a factional cluster-of-tastes regime.

Beyond validating the dataset itself, TASTE establishes a benchmark for criterion-level evaluation of design generation systems. Its multidimensional structure enables more precise comparison of model capabilities, supports routing and orchestration decisions based on task-specific design requirements, and provides supervision for preference judges, reward models, and alignment objectives for graphic-design generation. The data are available at \url{https://huggingface.co/datasets/purvanshi/TASTE} and the analysis pipeline at \url{https://github.com/purvanshi-lica/taste}; we recommend applying the same signal-validation protocol to future preference datasets before model training begins.

\bibliography{egbib}

@misc{Hirsch2026LICA,
  author       = {Elad Hirsch and Shubham Yadav and Mohit Garg and
                  Purvanshi Mehta},
  title        = {{LICA}: Layered Image Composition Annotations for Graphic
                  Design Research},
  howpublished = {arXiv preprint arXiv:2603.16098},
  year         = {2026},
}

@misc{Deganutti2026GDB,
  author       = {Adrienne Deganutti and Elad Hirsch and Haonan Zhu and
                  Jaejung Seol and Purvanshi Mehta},
  title        = {{Graphic-Design-Bench}: A Comprehensive Benchmark for
                  Evaluating {AI} on Graphic Design Tasks},
  howpublished = {arXiv preprint arXiv:2604.04192},
  year         = {2026},
}

@inproceedings{Bylinskii2017VisualImportance,
  author    = {Zoya Bylinskii and Nam Wook Kim and Peter O'Donovan and
               Sami Alsheikh and Spandan Madan and Hanspeter Pfister and
               Fredo Durand and Bryan Russell and Aaron Hertzmann},
  title     = {Learning Visual Importance for Graphic Designs and Data
                Visualizations},
  booktitle = {Proceedings of the 30th Annual ACM Symposium on User
                Interface Software and Technology (UIST)},
  year      = {2017},
}

@misc{Fosco2020Predimportance,
  author       = {Camilo Fosco and Vincent Casser and Amish Kumar Bedi and
                  Peter O'Donovan and Aaron Hertzmann and Zoya Bylinskii},
  title        = {Predicting Visual Importance Across Graphic Design Types},
  howpublished = {arXiv preprint arXiv:2008.02912},
  year         = {2020},
}

@misc{Survey2025DesignAI,
  author       = {Xingxing Zou and Wen Zhang and Nanxuan Zhao},
  title        = {From Fragment to One Piece: A Survey on {AI}-Driven
                  Graphic Design},
  howpublished = {arXiv preprint arXiv:2503.18641},
  year         = {2025},
}

@inproceedings{Huang2024VBench,
  author    = {Ziqi Huang and Yinan He and Jiashuo Yu and Fan Zhang and
               Chenyang Si and Yuming Jiang and Yuanhan Zhang and Tianxing
               Wu and Qingyang Jin and Nattapol Chanpaisit and Yaohui Wang
               and Xinyuan Chen and Limin Wang and Dahua Lin and Yu Qiao
               and Ziwei Liu},
  title     = {{VBench}: Comprehensive Benchmark Suite for Video Generative
                Models},
  booktitle = {Proceedings of the IEEE/CVF Conference on Computer Vision and
                Pattern Recognition (CVPR)},
  year      = {2024},
}

@misc{Huang2025VBench2,
  author       = {Dian Zheng and Ziqi Huang and Hongbo Liu and Kai Wang and
                  Jingwen He and Fan Zhang and Yuanhan Zhang and Jingkang
                  Yang and Yu Qiao and Ziwei Liu},
  title        = {{VBench-2.0}: Advancing Video Generation Benchmark Suite
                  for Intrinsic Faithfulness},
  howpublished = {arXiv preprint arXiv:2503.21755},
  year         = {2025},
}

@misc{Sun2026ImagenWorld,
  author       = {Tao Sun and others},
  title        = {{ImagenWorld}: Stress-Testing Image Generation Models with
                  Explainable Human Evaluation on Open-ended Real-World
                  Tasks},
  howpublished = {arXiv preprint arXiv:2603.27862},
  year         = {2026},
}

@misc{Ecomiq2026,
  author       = {Meiqi Sun and Mingyu Li and Junxiong Zhu},
  title        = {{E-comIQ-ZH}: A Human-Aligned Dataset and Benchmark for
                  Fine-Grained Evaluation of {E-commerce} Posters with
                  Chain-of-Thought},
  howpublished = {arXiv preprint arXiv:2602.21698},
  year         = {2026},
  note         = {CVPR 2026},
}

@inproceedings{Lim2025IHallA,
  author    = {Sang-gil Lim and Heesoo Jung and Choonghan Kim and
               Hyunwoo Park and Hwanhee Lee and Pilsung Kang},
  title     = {{I-HallA}: Evaluating Image Hallucination in Text-to-Image
                Generation with Question Answering},
  booktitle = {Proceedings of the AAAI Conference on Artificial Intelligence
                (AAAI)},
  year      = {2025},
}

@misc{Liu2025HallucinationSurvey,
  author       = {Hao Liu and others},
  title        = {A Survey of Multimodal Hallucination Evaluation and
                  Detection},
  howpublished = {arXiv preprint arXiv:2507.19024},
  year         = {2025},
}

@misc{Wu2023HPSv2,
  author       = {Xiaoshi Wu and Yiming Hao and Keqiang Sun and Yixiong Chen
                  and Feng Zhu and Rui Zhao and Hongsheng Li},
  title        = {{Human Preference Score v2}: A Solid Benchmark for
                  Evaluating Human Preferences of Text-to-Image Synthesis},
  howpublished = {arXiv preprint arXiv:2306.09341},
  year         = {2023},
}

@inproceedings{Ma2025HPSv3,
  author    = {Yuhang Ma and Xiaoshi Wu and Keqiang Sun and Hongsheng Li
               and others},
  title     = {{HPSv3}: Towards Wide-Spectrum Human Preference Score},
  booktitle = {Proceedings of the IEEE/CVF International Conference on
               Computer Vision (ICCV)},
  year      = {2025},
}

@inproceedings{Kirstain2023PickaPic,
  author    = {Yuval Kirstain and Adam Polyak and Uriel Singer and
               Shahbuland Matiana and Joe Penna and Omer Levy},
  title     = {{Pick-a-Pic}: An Open Dataset of User Preferences for
                Text-to-Image Generation},
  booktitle = {Advances in Neural Information Processing Systems (NeurIPS)},
  year      = {2023},
}

@inproceedings{Xu2023ImageReward,
  author    = {Jiazheng Xu and Xiao Liu and Yuchen Wu and Yuxuan Tong
               and Qinkai Li and Ming Ding and Jie Tang and Yuxiao Dong},
  title     = {{ImageReward}: Learning and Evaluating Human Preferences for
                Text-to-Image Generation},
  booktitle = {Advances in Neural Information Processing Systems (NeurIPS)},
  year      = {2023},
}

@inproceedings{Liang2024RichHF,
  author    = {Youwei Liang and Junfeng He and Gang Li and Peizhao Li and
               Arseniy Klimovskiy and Nicholas Carolan and Jiao Sun and
               Jordi Pont-Tuset and Sarah Young and Feng Yang and others},
  title     = {Rich Human Feedback for Text-to-Image Generation},
  booktitle = {Proceedings of the IEEE/CVF Conference on Computer Vision and
               Pattern Recognition (CVPR)},
  year      = {2024},
}

@inproceedings{Zhang2024MPS,
  author    = {Sixian Zhang and Bohan Wang and Junqiang Wu and Yan Li and
               Tingting Gao and Di Zhang and Zhongyuan Wang},
  title     = {Learning Multi-Dimensional Human Preference for Text-to-Image
               Generation},
  booktitle = {Proceedings of the IEEE/CVF Conference on Computer Vision and
               Pattern Recognition (CVPR)},
  year      = {2024},
}

@inproceedings{Xu2026VisionReward,
  author    = {Jiazheng Xu and Yu Huang and Jiale Cheng and Yuanming Yang
               and Jiajun Xu and Yuan Wang and Wenbo Duan and Shen Yang and
               Qinkai Li and Mingyi Zhang and others},
  title     = {{VisionReward}: Fine-Grained Multi-Dimensional Human
                Preference Learning for Image and Video Generation},
  booktitle = {Proceedings of the AAAI Conference on Artificial Intelligence
                (AAAI)},
  year      = {2026},
}

@misc{Peng2025DesignPref,
  author       = {Yi-Hao Peng and Jeffrey P. Bigham and Jason Wu},
  title        = {{DesignPref}: Capturing Personal Preferences in Visual
                  Design Generation},
  howpublished = {arXiv preprint arXiv:2511.20513},
  year         = {2025},
}

@inproceedings{Lin2024GenAIBench,
  author    = {Baiqi Li and Zhiqiu Lin and Deepak Pathak and Jiayao Li and
               Yixin Fei and Kewen Wu and Tiffany Ling and Xide Xia and
               Pengchuan Zhang and Graham Neubig and Deva Ramanan},
  title     = {{GenAI-Bench}: Evaluating and Improving Compositional
                Text-to-Visual Generation},
  booktitle = {Advances in Neural Information Processing Systems (NeurIPS)
                Datasets and Benchmarks Track},
  year      = {2024},
}

@article{Li2023AGIQA3K,
  author  = {Chunyi Li and Zicheng Zhang and Haoning Wu and Wei Sun and
             Xiongkuo Min and Xiaohong Liu and Guangtao Zhai and Weisi Lin},
  title   = {{AGIQA-3K}: An Open Database for {AI}-Generated Image Quality
              Assessment},
  journal = {IEEE Transactions on Circuits and Systems for Video Technology},
  year    = {2023},
}

@inproceedings{Li2024AIGIQA20K,
  author    = {Chunyi Li and Tengchuan Kou and Yixuan Gao and Yuqin Cao and
               Wei Sun and Zicheng Zhang and Yingjie Zhou and Zhichao Zhang
               and Weixia Zhang and Haoning Wu and Xiaohong Liu and
               Xiongkuo Min and Guangtao Zhai},
  title     = {{AIGIQA-20K}: A Large Database for {AI}-Generated Image
                Quality Assessment},
  booktitle = {CVPR Workshops (NTIRE)},
  year      = {2024},
}

@article{Chen2024AGIN,
  author  = {Zijian Chen and Wei Sun and Yuan Tian and Jun Jia and
             Zicheng Zhang and Jiarui Wang and Ru Huang and Xiongkuo Min
             and Guangtao Zhai and Wenjun Zhang},
  title   = {Exploring the Naturalness of {AI}-Generated Images},
  journal = {IEEE Transactions on Circuits and Systems for Video Technology},
  year    = {2024},
}

@misc{BlackForest2024Flux,
  author       = {{Black Forest Labs}},
  title        = {{FLUX.2}: Frontier Visual Intelligence},
  howpublished = {Model release, Black Forest Labs},
  year         = {2025},
  note         = {FLUX.2~[max]; released Dec.\ 2025. \url{https://bfl.ai}},
}

@misc{OpenAI2024DALLE,
  author       = {{OpenAI}},
  title        = {{GPT Image 1.5}},
  howpublished = {Model release, OpenAI},
  year         = {2025},
  note         = {Released Dec.\ 2025.},
}

@misc{Google2024Imagen,
  author       = {{Google DeepMind}},
  title        = {{Nano Banana 2}: {Gemini} Image Generation},
  howpublished = {Model release, Google DeepMind},
  year         = {2026},
  note         = {Released Feb.\ 2026.},
}

@misc{Bytedance2024Seedream,
  author       = {{ByteDance Seed}},
  title        = {{Seedream 5.0}},
  howpublished = {Model release, ByteDance Seed},
  year         = {2026},
  note         = {Seedream 5.0~Lite; released Feb.\ 2026.},
}

@techreport{Kamishima2003Sushi,
  author      = {Toshihiro Kamishima},
  title       = {Nantonac Collaborative Filtering: Recommendation Based on
                 Order Responses},
  institution = {ACM SIGKDD},
  year        = {2003},
  note        = {Sushi preference dataset.},
}

@article{Harper2015MovieLens,
  author  = {F. Maxwell Harper and Joseph A. Konstan},
  title   = {The {MovieLens} Datasets: History and Context},
  journal = {ACM Transactions on Interactive Intelligent Systems (TiiS)},
  volume  = {5},
  number  = {4},
  pages   = {19:1--19:19},
  year    = {2015},
}

@article{Mallows1957,
  author  = {C. L. Mallows},
  title   = {Non-Null Ranking Models},
  journal = {Biometrika},
  volume  = {44},
  number  = {1/2},
  pages   = {114--130},
  year    = {1957},
}

@article{BradleyTerry1952,
  author  = {Ralph Allan Bradley and Milton E. Terry},
  title   = {Rank Analysis of Incomplete Block Designs: I. The Method of
              Paired Comparisons},
  journal = {Biometrika},
  volume  = {39},
  number  = {3/4},
  pages   = {324--345},
  year    = {1952},
}

@article{Plackett1975,
  author  = {Robin L. Plackett},
  title   = {The Analysis of Permutations},
  journal = {Journal of the Royal Statistical Society: Series C},
  volume  = {24},
  number  = {2},
  pages   = {193--202},
  year    = {1975},
}

@book{Luce1959,
  author    = {R. Duncan Luce},
  title     = {Individual Choice Behavior: A Theoretical Analysis},
  publisher = {John Wiley \& Sons},
  year      = {1959},
}

@article{Kendall1938,
  author  = {Maurice G. Kendall},
  title   = {A New Measure of Rank Correlation},
  journal = {Biometrika},
  volume  = {30},
  number  = {1/2},
  pages   = {81--93},
  year    = {1938},
}

@book{Krippendorff2018,
  author    = {Klaus Krippendorff},
  title     = {Content Analysis: An Introduction to Its Methodology},
  edition   = {4th},
  publisher = {SAGE},
  year      = {2018},
}

@article{Krippendorff2004alpha,
  author  = {Klaus Krippendorff},
  title   = {Reliability in Content Analysis: Some Common Misconceptions
              and Recommendations},
  journal = {Human Communication Research},
  volume  = {30},
  number  = {3},
  pages   = {411--433},
  year    = {2004},
}

@inproceedings{Rafailov2023DPO,
  author    = {Rafael Rafailov and Archit Sharma and Eric Mitchell and
               Christopher D. Manning and Stefano Ermon and Chelsea Finn},
  title     = {Direct Preference Optimization: Your Language Model is
                Secretly a Reward Model},
  booktitle = {Advances in Neural Information Processing Systems (NeurIPS)},
  year      = {2023},
}

@article{qwen3vlembedding,
  title={Qwen3-VL-Embedding and Qwen3-VL-Reranker: A Unified Framework for State-of-the-Art Multimodal Retrieval and Ranking},
  author={Li, Mingxin and Zhang, Yanzhao and Long, Dingkun and Chen, Keqin and Song, Sibo and Bai, Shuai and Yang, Zhibo and Xie, Pengjun and Yang, An and Liu, Dayiheng and Zhou, Jingren and Lin, Junyang},
  journal={arXiv},
  year={2026}
}

@inproceedings{Zheng2023MTBench,
  author    = {Lianmin Zheng and Wei-Lin Chiang and Ying Sheng and
               Siyuan Zhuang and Zhanghao Wu and Yonghao Zhuang and
               Zi Lin and Zhuohan Li and Dacheng Li and Eric P. Xing and
               Hao Zhang and Joseph E. Gonzalez and Ion Stoica},
  title     = {Judging {LLM}-as-a-Judge with {MT-Bench} and {Chatbot Arena}},
  booktitle = {Advances in Neural Information Processing Systems (NeurIPS)
               Datasets and Benchmarks Track},
  year      = {2023},
}

@inproceedings{Tian2025MultiImageBias,
  author    = {Yu Tian and Tianqi Liu and Zhiyuan Liu and Jie Yang and
               Cordelia Schmid},
  title     = {Identifying and Mitigating Position Bias of Multi-image
                Vision-Language Models},
  booktitle = {Proceedings of the IEEE/CVF Conference on Computer Vision
                and Pattern Recognition (CVPR)},
  year      = {2025},
  note      = {arXiv:2503.13792},
}

@article{Goyal2026DistortBench,
  author    = {Divyanshu Goyal and Akhil Eppa and Vanya Bannihatti Kumar},
  title     = {{DistortBench}: Benchmarking Vision Language Models on
                Image Distortion Identification},
  journal   = {arXiv preprint arXiv:2604.19966},
  year      = {2026},
}

\appendix
\section{Statistical definitions}
\label{app:stats}

This appendix collects the formal definitions, support sets, and null
PMFs of the three signal-test statistics introduced in
\S\ref{sec:stats}.  Notation: a sample is one prompt rated by $R=5$
evaluators on $p=4$ items.  Each evaluator $r$ produces a strict
ranking $\pi_r : \{1,\dots,p\} \to \{1,\dots,p\}$.  Let $a_i, b_i$
($i=1,\dots,\binom{p}{2}$) enumerate item pairs with $a_i < b_i$.

\begin{table}[h]
\small
\centering
\begin{tabular}{p{2.4cm}p{4.6cm}p{6.0cm}}
\toprule
Statistic & Formula & Null support / PMF at $(p=4, R=5)$ \\
\midrule
Pairwise Kendall $\tau$
  & $\tau(\pi_r, \pi_s) = \dfrac{C_{rs} - D_{rs}}{\binom{p}{2}}$
  & Mahonian on $\{-1,\, -\tfrac{2}{3},\, -\tfrac{1}{3},\, 0,\, \tfrac{1}{3},\, \tfrac{2}{3},\, 1\}$ \\
\addlinespace
Per-prompt $T$
  & $T = \dfrac{1}{\binom{R}{2}} \sum_{r<s} \tau(\pi_r, \pi_s)$
  & Exact PMF on a finite support, mean $0$ \\
\addlinespace
Majority-vote prob.\ $\pmax$
  & $\pmax = \max\!\left(\dfrac{k}{R},\, 1 - \dfrac{k}{R}\right)$
  & $\{\tfrac{3}{5}, \tfrac{4}{5}, \tfrac{5}{5}\}$, PMF $(\tfrac{20}{32}, \tfrac{10}{32}, \tfrac{2}{32})$ \\
\addlinespace
Cycle indicator
  & $\mathbf{1}[\exists \text{ intransitive triple}]$
  & Bernoulli, null rate $\approx 0.211$ (MC) \\
\bottomrule
\end{tabular}
\caption{Signal-test statistics.  $C_{rs}$ and $D_{rs}$ count
concordant and discordant item-pairs between rankings $\pi_r$ and
$\pi_s$.  $k$ counts raters placing item $a$ above item $b$ for a
given pair.  The $\tau$ null is the Mahonian distribution; the $T$
null is its finite-mixture analogue; the $\pmax$ null is
symmetric-folded $\mathrm{Bin}(5, 1/2)$; the cycle null is computed
by Monte Carlo on $200{,}000$ samples.}
\label{tab:stat_defs}
\end{table}

\textbf{Goodness-of-fit tests.}  For each TASTE sub-dimension we
test the empirical histograms of $T$ and $\pmax$ against their null
PMFs using a chi-squared GOF test with low-count bin pooling
(adjacent bins are merged when expected count is below $5$).  The
cycle rate is tested with a one-sample binomial test against the
Monte-Carlo null rate of $0.211$.

\textbf{Subsampling reference anchors to $(p=4, R=5)$.}  Sushi has
$p=10$ items and $R \approx 5{,}000$ users; MovieLens uses
$1$-to-$5$ scalar ratings on a large movie pool; the HPSv2 test
split uses $p=10$ images (9 generative + 1 COCO real) per prompt
with $R=10$ fixed annotators.  For each anchor we draw items and
raters at random to match TASTE's shape (excluding the COCO real
image and restricting to the four strongest generators in the
HPSv2 case as described in \S\ref{sec:anchors}), feed the resulting
$5 \times 4$ rank matrix through the same statistic computations,
and pool many such draws to form a reference distribution.

\section{VLM-judge supplementary tables and methodology}
\label{app:vlmjudge}

This appendix supplements \S\ref{sec:res:vlmjudge}: per-VLM
diagnostic metrics, per-criterion position-bias rates, cross-VLM
correlations, and full protocol details.

\subsection{VLM diagnostics: position-bias, conditional accuracy, and paraphrase stability}

\begin{table}[h]
\centering
\small
\setlength{\tabcolsep}{4pt}
\begin{tabular}{lrrrr}
\toprule
Model & Macro acc. & Pos-bias & Cond. acc. & Para $\sigma$ \\
\midrule
Qwen3-VL-4B-Instruct    & 0.530 & 0.775 & 0.630 & 0.055 \\
Qwen3-VL-8B-Instruct    & 0.539 & 0.679 & 0.613 & 0.060 \\
Qwen3-VL-32B-Instruct   & 0.536 & 0.463 & 0.567 & 0.064 \\
Gemma-3-27B-it          & 0.528 & 0.439 & 0.546 & 0.065 \\
Kimi-VL-A3B-Instruct    & 0.509 & 0.868 & 0.656 & 0.054 \\
InternVL3.5-14B         & 0.525 & 0.551 & 0.550 & 0.057 \\
\bottomrule
\end{tabular}
\caption{VLM-only diagnostics for
Table~\ref{tab:vlm-baselines}.  \textbf{Pos-bias} is the fraction
of pairs whose verdict is unchanged when image order is flipped
(\ie, determined by position rather than content).  \textbf{Cond.\
acc.}\ is agreement with the designer majority on the
order-consistent fraction of pairs only.  \textbf{Para $\sigma$}
is the standard deviation of per-criterion accuracy across the
eight question paraphrases.}
\label{tab:vlm-diag}
\end{table}

\FloatBarrier
\subsection{Per-criterion position-bias rate}

\begin{table}[h]
\centering
\footnotesize
\setlength{\tabcolsep}{4pt}
\begin{tabular}{lrrrrrr}
\toprule
Criterion & Q-4B & Q-8B & Q-32B & G-27B & K-3B & I-14B \\
\midrule
aesthetics\_color\_harmony & 0.900 & 0.779 & 0.463 & 0.367 & 0.910 & 0.581 \\
aesthetics\_mood           & 0.825 & 0.708 & 0.398 & 0.346 & 0.875 & 0.493 \\
aesthetics\_preference     & 0.529 & 0.585 & 0.281 & 0.315 & 0.792 & 0.367 \\
aesthetics\_typography     & 0.896 & 0.750 & 0.410 & 0.440 & 0.913 & 0.598 \\
aesthetics\_visual\_hier   & 0.890 & 0.773 & 0.431 & 0.537 & 0.927 & 0.629 \\
descriptions\_color\_acc   & 0.735 & 0.637 & 0.658 & 0.542 & 0.860 & 0.621 \\
descriptions\_preference   & 0.660 & 0.592 & 0.406 & 0.440 & 0.846 & 0.498 \\
descriptions\_spatial\_acc & 0.733 & 0.673 & 0.652 & 0.498 & 0.860 & 0.602 \\
descriptions\_typography   & 0.810 & 0.613 & 0.471 & 0.467 & 0.825 & 0.567 \\
\bottomrule
\end{tabular}
\caption{Per-criterion position-bias rate for the six VLM judges.
Aesthetics-side criteria that demand more abstract judgment (color
harmony, visual hierarchy, mood) consistently show the highest
position bias for every model; preference criteria show the
lowest.}
\label{tab:vlm-percrit-posbias}
\end{table}

\FloatBarrier
\subsection{Cross-VLM correlation statistics}

\begin{table}[h]
\centering
\small
\begin{tabular}{lrc}
\toprule
Variable pair & $\rho$ & $p$ \\
\midrule
Pos-bias vs Cond.\ acc.\        & $+0.94$ & $0.005$ \\
Pos-bias vs Macro acc.\         & $-0.26$ & $0.62$~(\textit{n.s.}) \\
Cond.\ acc.\ vs Macro acc.\     & $-0.09$ & $0.87$~(\textit{n.s.}) \\
\bottomrule
\end{tabular}
\caption{Spearman correlations across the six open-weight VLMs in
the slate.  Position-bias rate covaries with conditional accuracy
(rank-correlated $\rho{=}{+}0.94$, $p{=}0.005$); neither
separately predicts macro accuracy.}
\label{tab:vlm-spearman}
\end{table}

\FloatBarrier
\subsection{Protocol details}

Each VLM is queried with greedy (argmax) decoding to obtain a
single deterministic verdict per query.  Each VLM is run twice on
every pair, in both image orders ($8{,}640$ inferences per model),
and a verdict is scored under MT-Bench
S1~\cite{Zheng2023MTBench}: weight $1$ if it agrees with the
designer majority \emph{and} is consistent across both orderings,
weight $0$ if it disagrees and is consistent, and weight $0.5$ if
it is order-inconsistent.

The text prompt has four parts: (i) the brief shown to designers,
verbatim; (ii) the criterion name; (iii) a one-paragraph rubric
that summarises the criterion (a one-sentence rephrasing of the
guidance designers received during pilot interviews); and (iv) a
question.  The brief portion is byte-identical to the text shown to
designers in the annotation interface, so the VLM and the
designers see the same description of the design task.  To control
for phrasing sensitivity, the question is drawn from a pool of
eight per-criterion paraphrases, deterministically assigned per
$(\text{prompt\_id}, \text{model}_A, \text{model}_B)$ tuple by
hash; the per-paraphrase accuracy standard deviation is reported
as ``Para $\sigma$'' in Table~\ref{tab:vlm-diag}.  A worked example
for the descriptions-typography criterion is below; the bracketed
ellipsis stands for the brief paragraphs.

\begin{quote}
\footnotesize\ttfamily
You are a professional graphic designer evaluating two AI-generated
images for the same brief.\\[2pt]
BRIEF: [User Intent + Description, verbatim from the designer
view]\\[2pt]
CRITERION: Typography (Descriptions)\\
RUBRIC: Whether any text demanded by the prompt is rendered
correctly. Consider whether the requested words appear, whether
they are spelled correctly, and whether typography choices fit the
brief's directions.\\[2pt]
Image A: [image A]\quad Image B: [image B]\\[2pt]
Which image more accurately renders the text demanded by the
prompt? Reply with a JSON code block containing the answer letter.
\end{quote}

The dedicated scorers are evaluated once per pair with only the
User Intent paragraph as the text input, matching the noun-phrase
distribution of the prompts they were trained on.  All inputs and
verdicts are released alongside the dataset.

\section{Designer evaluation instructions (excerpt)}
\label{app:guidelines}

For reference, the following is the opening context paragraph shown
to designers at the start of every annotation session.  It defines
the task framing, the input and output structure, and the standard
designers were asked to apply.  This text is reproduced verbatim
from the live annotation interface.

\begin{quote}
\itshape
This evaluation focuses on AI-generated graphic design assets.  You
will be reviewing outputs from multiple AI models, each generating
designs like logos, posters, flyers, social media graphics, and
print materials from the same text prompt.

\textbf{Input:} Each task will include a prompt that combines the
user's intent with either a detailed description of the layout or
an aesthetic direction.  You will see one prompt per task and will
be assigned an equal number of each prompt type.  You will never
see both prompts from the same layout.

\textbf{Output:} Each AI model generates a single design asset
based on that prompt.

These outputs serve a clear creative purpose: they need to function
as real, usable design work.  These should be the kind of asset a
brand would actually post, print, or ship.  As a professional
designer, you already know what separates a layout that
communicates from one that just fills space.  You know when type
hierarchy works, when visual style matches the brief, and when
something looks ``close'' but still feels off.  That eye is exactly
what this evaluation needs.
\end{quote}

\section{Annotator Profiles}
\label{app:annotators}

TASTE was annotated by ten professional graphic designers recruited through Contra, a professional network of creative practitioners. Annotators were selected based on demonstrated commercial design experience, portfolio quality, and prior screening through the Contra Labs Network. Collectively, the cohort spans brand identity design, visual systems, web and product design, performance marketing creative, illustration, print design, and presentation design.

The annotators represent a broad cross-section of contemporary design practice, including specialists in brand systems, Framer-based web design and development, conversion-oriented advertising creative, and editorial / illustrative design. Several participants had extensive commercial track records on Contra (ranging from 20+ to 120+ paid projects), while others were selected based on strong portfolio quality despite shorter platform histories. Across the cohort, designers reported experience delivering work for enterprise clients, venture-backed startups, consumer brands, and nonprofit organizations.

Table~\ref{tab:annotator-profiles} summarizes the professional backgrounds of the annotator cohort.

\begin{table*}[t]
\centering
\small
\resizebox{\textwidth}{!}{
\begin{tabular}{p{3.2cm}p{3.8cm}p{10cm}}
\toprule
\textbf{Primary Skill} & \textbf{Secondary Skill(s)} & \textbf{Professional Background} \\
\midrule
Brand Identity Design & Design Systems, Visual Design & High-volume brand identity practice with 80+ commercial projects spanning productized brand sprints and long-term retainers. \\

Brand Identity Design & Brand Strategy & Brand identity specialist focused on rapid identity-development engagements for commercial clients. \\

Framer Web Design \& Development & Brand Identity Design & Senior web and brand designer specializing in Framer-based website development and design-to-production workflows. \\

Brand \& Product Design & Framer Web Development, UX Design & Multidisciplinary designer with extensive experience spanning brand systems and interactive product surfaces. \\

Graphic Design & Brand Identity, Email Marketing Design & Designer focused on brand systems, social creative, and lifecycle marketing assets. \\

Brand Identity Design & Framer Web Design & Branding and web-design specialist working across identity systems and web implementation. \\

Brand Identity Design & Web Development & Hybrid brand designer / developer delivering integrated identity and web systems. \\

Graphic Design \& Illustration & Brand Identity, Print \& Presentation Design & Designer with agency experience spanning illustration, print communication, and presentation design. \\

Performance Ad Design & Brand Identity, Email Marketing & Marketing designer specializing in static ad creative and conversion-oriented campaign assets. \\

Brand Identity \& Logo Design & Illustration & Graphic designer focused on logo systems, brand marks, and illustrative identity work. \\
\bottomrule
\end{tabular}}
\caption{Professional backgrounds of the ten TASTE annotators.}
\label{tab:annotator-profiles}
\end{table*}

\end{document}